\documentclass[10pt,twocolumn,letterpaper]{article}

\usepackage{cvpr}              
\usepackage[accsupp]{axessibility} 
\usepackage[dvipsnames]{xcolor}

\definecolor{cvprblue}{rgb}{0.21,0.49,0.74}
\usepackage[pagebackref,breaklinks,colorlinks,citecolor=cvprblue]{hyperref}
\usepackage{makecell}
\usepackage[flushleft]{threeparttable}
\usepackage{caption}
\usepackage{multirow}
\usepackage{arydshln}
\usepackage{amsmath}
\usepackage{colortbl}

\def\paperID{7593} 
\def\confName{CVPR}
\def\confYear{2024}

\title{
Benchmarking Implicit Neural Representation and Geometric Rendering in Real-Time RGB-D SLAM}

\author{Tongyan Hua$^{1}$ \quad Lin Wang$^{1}$$^{,2}$\thanks{Corresponding author.}\\
$^{1}$AI Thrust, HKUST(GZ)  \quad $^{2}$Dept. of CSE, HKUST\\
{\tt\small  t.hua.msc@outlook.com, linwang@ust.hk}
\\
\small{Project Page: \url{https://vlis2022.github.io/nerf-slam-benchmark/}}
}

\begin{document}
\maketitle

\begin{abstract}

Implicit neural representation (INR), in combination with geometric rendering, has recently been employed in real-time dense RGB-D SLAM.
Despite active research endeavors being made, there lacks a unified protocol for fair evaluation, impeding the
evolution of this area. In this work, we establish, to our knowledge, the \textbf{first} open-source benchmark framework to
evaluate the performance of a wide spectrum of commonly used INRs and rendering functions for mapping and localization. The goal of our benchmark is to 1) gain an intuition of how different INRs and rendering functions impact mapping and localization and 2) establish a unified evaluation protocol \wrt. the design choices that may impact the mapping and localization. With the framework, we conduct a large suite of experiments, offering various insights in choosing the INRs and geometric
rendering functions: for example, the dense feature grid outperforms other INRs (\eg tri-plane and hash grid), even when geometric and color features are jointly encoded for memory efficiency. To extend the findings into the practical scenario, a hybrid encoding strategy is proposed to bring the best of the accuracy and completion from the grid-based and decomposition-based INRs. We further propose explicit hybrid encoding for high-fidelity dense grid mapping to comply with the RGB-D SLAM system that puts the premise on robustness and computation efficiency. 

\end{abstract}

\begin{figure}[t]
  \centering
   \includegraphics[width=1\linewidth]{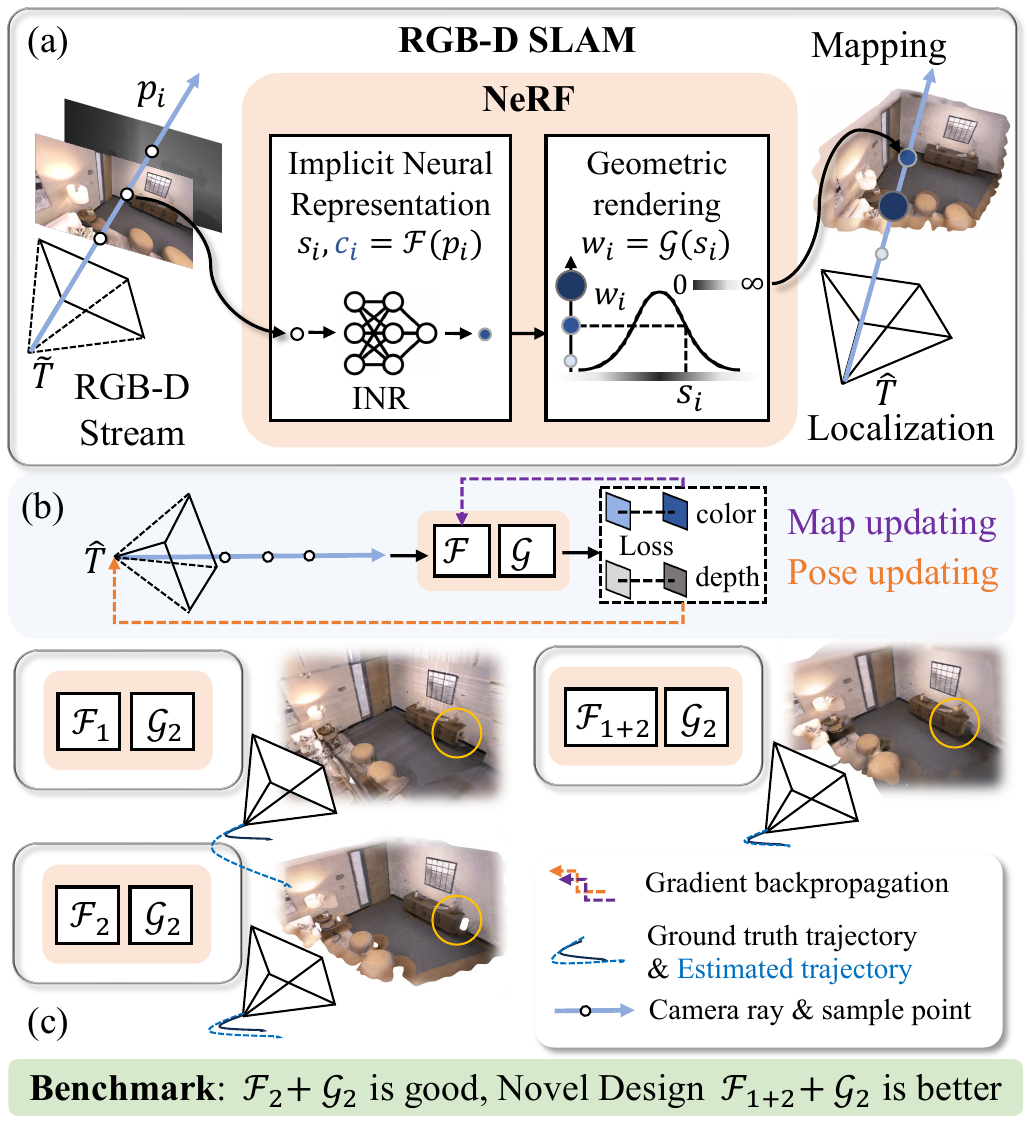}
   \vspace{-18pt}
   \caption{(a) We establish a novel benchmark to evaluate different elements of NeRF, narrowly defined as a combination of INR function \textbf{$\mathcal{F}$} and geometric rendering function \textbf{$\mathcal{G}$}, under the unified RGB-D SLAM paradigm.
   (b) Rendering Loss guides the online updating of the pose from $\tilde{T}$ to $\hat{T}$, and parameter of $\mathcal{F}$. 
(c) A toy example illustrates the impact of various combinations of \textbf{\(\mathcal{F}\)} and \textbf{\(\mathcal{G}\)}: \(\mathcal{F}_2\) surpasses \(\mathcal{F}_1\) in trajectory estimation and reconstruction fidelity but compromising completeness, inspire new designs that bring the benefits of \(\mathcal{F}_1\) and \(\mathcal{F}_2\) to form a hybrid encoding \textbf{\(\mathcal{F}_{1+2}\)}.}

      \label{fig:psudo1}
      \vspace{-10pt}
\end{figure}

\vspace{-15pt}
\section{Introduction}
\label{sec:intro}

Simultaneous Localization and Mapping (SLAM) is a pivotal task in 3D computer vision, with the goal of estimating the position and orientation of a sensor, while concurrently building a map of the surrounding scene. 
For the real-time dense visual SLAM system, a large number of methods have been proposed, predominantly based on the RGB-D cameras~\cite{RGBDSLAM1,RGBDSLAM2,RGBDSLAM3,RGBDSLAM4,RGBDSLAM5,RGBDSLAM6,RGBDSLAM7}.


Neural Radiance Field (NeRF) is an emerging technique based on the \textit{Implicit Neural Representation (INR)}, in combination with \textit{geometric rendering} for novel view synthesis~\cite{nerf21}. It employs a Multilayer Perceptron (MLP) to map a 3D point (\ie, spatial location along viewing direction) to density and color. 
Various NeRF variants have emerged since, featuring unique representations such as hash-grid~\cite{instant-npg22} and tri-plane~\cite{triplane21}, or focusing on mapping the 3D point to different geometric properties, such as the surface ~\cite{cole2021differentiable,UNISURFUN21,VolumeRO21,towards23,neus21}, while developing corresponding geometric rendering strategies. 
NeRF effectively models the intrinsic structure of a specific scene, capturing its geometry in a compact yet expressive manner that aligns closely with observed locations. As a result, it inherently incorporates the corresponding observation positions. This feature facilitates the deduction of the camera poses directly from the trained NeRF~\cite{pose0,pose1,pose2,pose3,pose4}.

This has inspired active research endeavors, integrating NeRF with RGB-D SLAM, demonstrated by the increasing number of publications \cite{imap22,niceslam22,voxfusion22,eslam23,coslam23,nerfslam22,orbeez23,goslam23,newton23,hislam23arXiv,fmapping23}. In general, these methods can be classified into: 1) NeRF-centric methods, where NeRFs are used for both scene reconstruction and pose estimation~\cite{imap22,niceslam22,voxfusion22,eslam23,coslam23}, as depicted in~\cref{fig:psudo1}(a) and (b), and 2) SLAM-centric methods, where the location is provided by other SLAM systems \cite{nerfslam22,orbeez23,goslam23,newton23,hislam23arXiv}.
Although tremendous efforts have been made to reconstruct high-fidelity scenes and improve pose estimation, several limitations persist with the active progress of research:

\textbf{L1: The Absence of a Unified and Comprehensive Benchmark Framework.} This hampers the comparison of different NeRFs within the RGB-D SLAM system. State-of-the-art (SOTA) NeRF-SLAM methods~\cite{imap22,niceslam22,voxfusion22,eslam23,coslam23,pointslam24} usually exhibit a variety of strategies regarding components 
other than the INR and rendering, 
\eg, how training data is selected (refer to as keyframe selection in SLAM~\cite{niceslam22}),
This makes it hardly possible to directly compare systems and capture the actual progress stemming from the NeRFs' design. Consequently, it is imperative to understand the individual characteristics of NeRF when designing the SLAM system for varied purposes.

\textbf{L2: Lack of Assessment of NeRF Component Variations on SLAM Performance.}
As depicted in~\cref{fig:psudo1}(c), NeRF, defined by \textbf{$\mathcal{F}$} and \textbf{$\mathcal{G}$}.
has many variants that can significantly affect the performance of SLAM systems.
The network architecture choice of \textbf{$\mathcal{F}$}, for example, is vital for learning scene representations accurately and efficiently. 
Some architectures, \eg, joint color and geometry encoding of Tri-plane, can quickly converge but may lose important details. This influences the precision of mapping and tracking within NeRF-enhanced SLAM systems. Moreover, the rendering methods used to integrate geometry and color information along rays are also critical, since the quality of rendered pixels directly influences the pose and map updating, as~\cref{fig:psudo1}(b).
While high-accuracy methods~\cite{UNISURFUN21,VolumeRO21,towards23,neus21} improve the rendering quality, they generally increase computational demands and may not perform well with less accurately estimated poses.

\begin{figure*}[t!]
\centering
\includegraphics[width=1\linewidth]{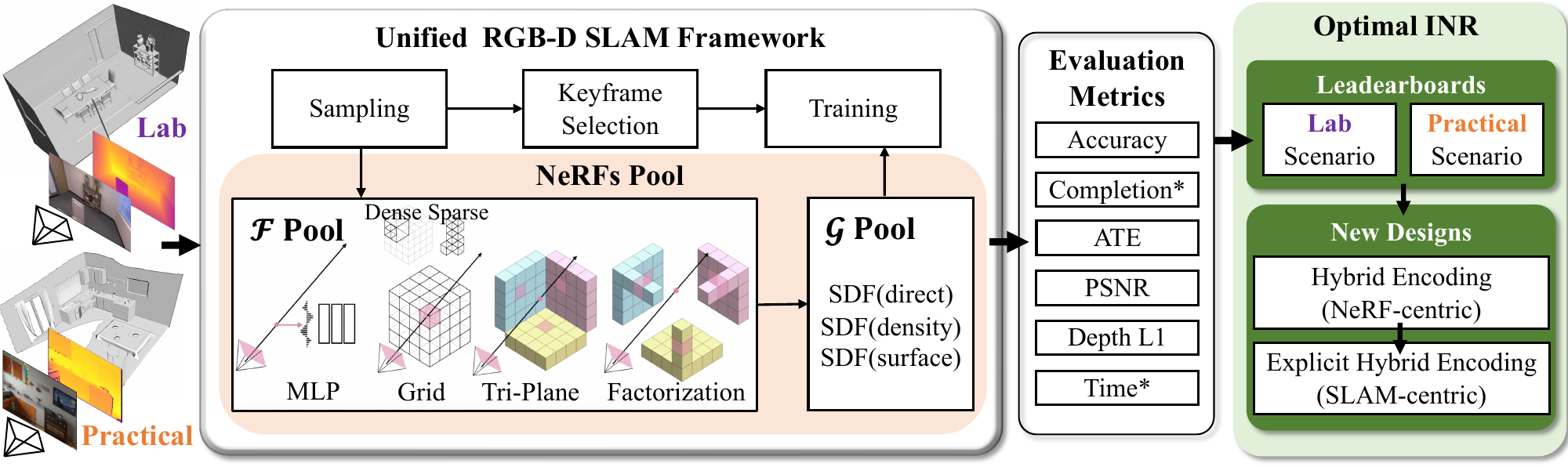}
    \vspace{-20pt}
   \caption{The proposed pipeline for NeRF-SLAM benchmark. The asterisk \mbox{*} indicates the existing two values for evaluation.}
   \label{fig:main_framework}
   \vspace{-12pt}
\end{figure*}

In this paper, we establish, to our knowledge, the \textbf{first} open-source benchmark framework to evaluate the performance of a wide spectrum of commonly used \textbf{$\mathcal{F}$} and \textbf{$\mathcal{G}$} for examining their effectiveness of mapping and localization. 
The major contributions are summarized as follows:

\textbf{C1: Comprehensive Evaluation of NeRFs within a Unified RGB-D SLAM Framework.} 
We propose a novel RGB-D SLAM benchmark framework, featuring a unified evaluation protocol to assess different NeRF components effectively.
We unfold our benchmark from a NeRF-centric paradigm. It covers five major variables that categorizing into two categories in~\cref{fig:main_framework}, \ie, the unified SLAM framework (including uniformed implemented sampling, training, and keyframe selection) and the NeRFs as a combination of $\mathcal{G}$ and $\mathcal{F}$. Our main objective is to investigate how the NeRFs influence the SLAM performance under uniformly controlled configurations (\cref{sub:standard-salm}) in two established scenarios: lab and practical scenarios (\cref{sub:scenario}). 

\textbf{C2: Pivotal Insights and Derived New Designs.} 
We reveal the significant differences in the performance of INRs in RGB-D SLAM problems attributable to their structural paradigms. Specifically, hybrid representations that simplify the 4D feature space, such as hash grid and tri-plane, often require separate encoding of geometry and appearance to achieve optimal performance, whereas representations of complete forms, such as dense grid and pure MLP, do not necessitate this. 
We also discovered that, in datasets with complete trajectory loops, \eg Replica dataset~\cite{replica19arxiv}, grid-based INRs (hash grid and dense grid) show better performance, while decomposition-based INRs (tri-plane and factorization) exhibit superior efficacy under random, loop-free trajectories, \eg the sequences in~\cite{neuralrgbd22}. This phenomenon inspired us to propose a novel blending of grid-based and decomposition-based methods.

\textbf{C3: Bags of Engineering Tricks for the Extension to Mapping.} 
Our research demonstrates that a strategic blend of dense grid representation, initially introduced in~\cite{niceslam22}, and appropriate geometric rendering functions can not only surpass more recently proposed sparse alternatives~\cite{coslam23}, but also surpass its application in earlier SLAM systems~\cite{niceslam22}. 
This finding suggests that our new SLAM framework successfully capitalizes on a suite of the latest engineering techniques.
Our benchmark also indicates that minimal feature dimensions are sufficient for achieving relatively high-quality mapping and tracking. This enables us to achieve extremely fine spatial partitioning without a substantial increase in memory usage. 
Based on this, we have effectively re-engineered the dense grid -- a scene representation known for its high memory demand but exceptional leaderboard performance -- for real-time high-fidelity mapping.

\section{Related Works}
\label{sec:related}

\textbf{NeRF-centric} approaches stem from NeRF's inherent potential for pose estimation~\cite{pose0,pose1,pose2,pose3,pose4}. The pioneering study~\cite{imap22} showcased that the foundational NeRF model~\cite{nerf21} could act as the sole representation for concurrent localization and mapping.
This spurred subsequent research that illustrated the advantages of hybrid representations~\cite{nicer23,voxfusion22,eslam23,coslam23} over singular MLP structures. Predominantly, dense SLAM methods have relied on RGB-D inputs to accelerate convergence on sampling distributions. Yet, recent efforts have ventured into dense SLAM with RGB-only inputs, where multi-level dense feature grids have yielded impressive results, whether an external depth estimator is used~\cite{niceslam22} or not~\cite{dimslam23}. Another development is the shift in volume density representation from occupancy~\cite{imap22,niceslam22,dimslam23,pointslam24} to Signed Distance Fields (SDF)~\cite{voxfusion22,eslam23,coslam23,nicer23}, with the latter demonstrating rapid convergence and superior reconstruction quality. Furthermore, these methods have laid the groundwork for semantic tasks, enhancing scene understanding and enabling real-time reconstruction~\cite{task_semantic23,task_vmap23}, and have even shown promise in style transfer applications~\cite{task_unifusion23}.

\textbf{SLAM-centric methods} were developed to leverage NeRF as an external module within a self-content SLAM system, aimed at 
achieving the robustness similar to well-acknowledged visual SLAM systems~\cite{tra_21orb3,tra_Engel2014LSDSLAMLD,tra_Leutenegger2015KeyframebasedVO,tra_MurArtal2015ORBSLAMAV,tra_MurArtal2016ORBSLAM2AO}.
Initial adoption of hash-based sparse parametric encoding~\cite{instant-npg22} was favored for its memory efficiency and quick convergence~\cite{nerfslam22,orbeez23}. 
Some recent studies have begun to re-explore the superiority of purely MLP-based spatial representations~\cite{multi-mlp23,eMLP23}.
The potential for dense reconstruction and effective point cloud compression has led to a focus on neural implicit mapping using posed RGB-D observations~\cite{shine-map23,nerf-loam23,rim23,H2Mapping23}, and even solely from posed RGB inputs~\cite{fmapping23} in contexts such as robotics and autonomous driving. Further, some research has expanded these methods for practical applications, addressing challenges in large-scale mapping and multi-robot mapping fusion~\cite{sbumap_map-fusion23,sbumap_mips23,sbumap_nisb23}.

\vspace{-8pt}
\section{NeRF-SLAM Benchmark}
\label{sec:benchmark}
\noindent\textbf{Problem Formulation.} Given a stream of synchronized RGB-D input frames ${I, D}_t$ at timestamp $t$, the color and depth of a pixel $x$ are represented by $c_x$ and $d_x$, respectively. Along the camera ray passing through pixel $x$, we sample $N$ points, each associated with a specific sample $p_i$ at a distance $d_i$. A learnable neural implicit function $F(\cdot)$ is then employed to predict the appearance $c_i$ and geometric properties $g_i$ of each sample:
{\setlength\abovedisplayskip{2pt}
\setlength\belowdisplayskip{2pt}
\begin{equation}
(c_i,g_i) = F(p_i)
  \label{eq:nerf_pre}
\end{equation}}To determine the weight $w_i$ for each sample along a ray, we employ a geometric rendering function $G(\cdot)$:
{\setlength\abovedisplayskip{2pt}
\setlength\belowdisplayskip{2pt}
\begin{equation}
w_i = G(g_i)
  \label{eq:nerf_geo}
\end{equation}}The color and depth can be estimated as:
{\setlength\abovedisplayskip{2pt}
\setlength\belowdisplayskip{2pt}
\begin{equation}
\tilde{c}_x=\sum^N_{i=1} w_i\cdotp c_i, \space
\tilde{d}_x=\sum^N_{i=1} w_i\cdotp d_i
  \label{eq:nerf_render}
\end{equation}}
%
We formulate NeRF-SLAM as a continuous online learning task. The training data, \ie, the sampled rays through the pixel $x$, are cached, for the continuous optimization of \( F(\cdot) \) and camera pose \( T_t \). This process adheres to the fundamental photometric \( e^p_x \) and geometric \( e^g_x \) constraints:
{\setlength\abovedisplayskip{2pt}
\setlength\belowdisplayskip{2pt}
\begin{equation}
e^p_x=\tilde{c}_x - c_x,
e^g_x=\tilde{d}_x - d_x
  \label{eq:nerf_loss}
\end{equation}}
\subsection{NeRFs for RGB-D SLAM}
\label{sub:nerf-method}
\tabcolsep=0.06cm
{\setlength\abovedisplayskip{2pt}
\setlength\belowdisplayskip{2pt}
\begin{table}[t!]
    \centering
    \begin{threeparttable}
    \scalebox{0.9}{
    \begin{tabular*}{0.45\textwidth}{cc}
    \toprule
        \textbf{$\mathcal{F}$} & $F_{geo} \in \mathcal{F}$\\
         \midrule
         MLP$^1$ &  $(g_i,h_i) =MLP(\gamma(p_i)).$\\
        \hdashline 
        \multirow{2}{*}{Grid$^2$} &  $(g_i,h_i)=MLP(\Phi_{dense}^{*}(p_i),\gamma(p_i)).$\\
            
            & $(g_i,  h_i ) =MLP(\Phi_{hash}^{*}(p_i),\gamma(p_i)).$\\
        \hdashline
         \multirow{2}{*}{Decomposition$^2$} & $(g_i,h_i) =MLP(\Phi_{tri}^{*}(p_i),\gamma(p_i)),$\\
        & $(g_i,h_i) =MLP(\Phi_{fac}^{*}(p_i),\gamma(p_i)).$\\
         \midrule
         \textbf{$\mathcal{G}$}&  $G \in \mathcal{G}$\\
         \midrule
         SDF (direct)$^4$ & $w_i=sig(\frac{g_i}{tr})\cdotp sig(-\frac{g_i}{tr}).$\\
        \hdashline
         \multirow{2}{*}{SDF (density)$^5$}& $w_i = exp(-\sum\limits_{k=1}^{i-1} \sigma_i) (1-exp(-\sigma_i), $\\
 & $\sigma_i=\beta\cdotp sig(-\beta\cdotp g_i).$\\
         \hdashline
         \multirow{2}{*}{SDF (surface)$^6$}& $w_i=\alpha_i\prod_{j=1}^{i-1}(1-\alpha_j), $\\
 & $\alpha_i=\max\left(\frac{sig(g_i)-{sig(g_{i+1})}}{sig(g_i)},0\right).$\\
    \bottomrule
    \end{tabular*}
    }
      \vspace{-8pt}
      \captionsetup{justification=justified}
    \caption{The selection of $F_{geo}(\cdot)$ and $G(\cdot)$. $^*$ denotes the existence of multi-resolution spatial splits, and $\gamma(\cdot)$ represents the positional encoding function, and $sig$ refers to sigmoid function.
    $\mathcal{F}^1$ represents a pure multi-layer perceptron~\cite{imap22};
    $dense$ and $hash$ in $\mathcal{F}^2$ refers to the dense~\cite{niceslam22} and hash~\cite{coslam23} feature grid encoding, respectively; $tri$ and $fac$ in $\mathcal{F}^3$ denote tri-plane~\cite{eslam23} and factorization~\cite{hua2024himap} encoding, respectively.
    %
    $tr$ in $\mathcal{G}^4$ stands for truncation of SDF~\cite{coslam23,voxfusion22}; 
    the $\beta$ in $\mathcal{G}^5$ stands a learnable parameter~\cite{eslam23}, and the $\mathcal{G}^6$ is originate from~\cite{neus21}and adopted by~\cite{goslam23}.}
    \label{tab:representation}
        \vspace{-10pt}
    \end{threeparttable}
\end{table}}

To address the RGB-D SLAM problem, our objective is to formulate the neural implicit function \( F(\cdot) \) and the geometric function \( G(\cdot) \). As such, they can efficiently approximate the true density distribution along a camera ray, thereby providing an accurate estimation of \( w_i \):
{\setlength\abovedisplayskip{2pt}
\setlength\belowdisplayskip{2pt}
\begin{equation}
w_i = G(F_{geo}(p_i)), \\
F \in \mathcal{F}, G \in \mathcal{G}
  \label{eq:weight}
\end{equation}}Denoting learnable geometric and appearance implicit neural functions \( \{ F_{\text{geo}}, F_{\text{app}}\}\in F  \), 
consolidating choices from prominent baselines~\cite{imap22,niceslam22,voxfusion22,eslam23,coslam23,goslam23}, we firstly unified the formulation of \( F_{\text{geo}} \) in~\cref{tab:representation}, laying the groundwork for defining \( G(\cdot) \).

A key benefit of employing the SDF over alternative representations, such as occupancy grids, lies in its capability to leverage per-point losses across all samples and losses from the rendered image. 
This contributes to the model's fast convergence, a finding validated by~\cite{eslam23,coslam23}. 
Despite occupancy grids' efficacy across various domains, particularly in robotics, this paper focuses on the high-fidelity reconstruction and accelerated convergence offered by SDFs. This focus aligns with that of the recent SOTA methods~\cite{eslam23,nicer23}. 
Thus, discussions on the occupancy are excluded, with $g_i$ interchangeably referred to as $s_i$.
In the following, we delineate three structural paradigms for the appearance functions corresponding to each \( \mathcal{F} \):
{\setlength\abovedisplayskip{2pt}
\setlength\belowdisplayskip{2pt}
\begin{equation}
c_i =
\begin{cases}
    MLP(\Phi^{*}(p_i),\gamma(p_i)), & \text{if } \text{ coupled(base)} \\
    MLP(\Phi^{*}(p_i),\gamma(p_i),h_i),& \text{if } \text{ coupled} \\
    MLP(\phi^{*}(p_i),\gamma(p_i)). & \text{if } \text{ decoupled} 
\label{eq:nerf_state}
\end{cases}
\end{equation}}The distinction between coupled and decoupled structures hinges on the encoding of color: it is either independently encoded by \( \phi \) or jointly encoded with geometry by \( \Phi \). We further categorize the coupled structure into two types based on the presence of channeled geometric features.

In examining hybrid representations that incorporate structural elements beyond the standard \( MLP \), we observe a diversity of configurations \wrt the resolution levels and feature dimensions. For example, some studies assign high-dimensional features to spatial partitions, \eg, 32 dimensions as in~\cite{niceslam22,eslam23} and 16 in~\cite{voxfusion22}, alongside relatively low-resolution levels (3 in~\cite{niceslam22}, 2 in~\cite{eslam23}, and 1 in~\cite{voxfusion22}). Conversely, other studies utilize extremely low-dimensional features (2 dimensions in~\cite{dimslam23,coslam23}) while significantly increasing the number of multi-resolution levels (6 in~\cite{dimslam23} and 16 in~\cite{coslam23}). Also, the settings for resolutions vary widely: the finest granularity for appearance embeddings is set at 16cm in~\cite{niceslam22} and 3cm in~\cite{eslam23}, with the broadest level at 200cm in~\cite{niceslam22} and 24cm in~\cite{eslam23}. In this work, we ensure a fair comparison of these representations by controlling the feature dimensions and the number of resolution levels. This minimizes the computational overhead (\ie, opting for 2 resolution levels and 2 feature dimensions in constructing the leaderboard). 
\textit{Detailed visual descriptions for the impacts of feature dimension are available in the supplmat}.


While the sampling strategy is an ineligible component in NeRF, this paper does \textit{not consider} it a variable for examination, due to the substantial simplification of the sampling process,  afforded by the incorporation of additional depth input.
Therefore, we leverage the off-the-shelf sampling strategy for a unified SLAM framework in the following section, based on the validated techniques proposed in the latest research~\cite{niceslam22,nicer23,imap22}.

\subsection{Unified Evaluation Protocol}
\label{sub:standard-salm}
In this section, we describe the proposed unified SLAM framework for evaluating the possible combinations, detailed in~\cref{tab:representation} and~\cref{eq:nerf_state}. The schematic diagram in~\cref{fig:main_framework} delineates the three primary components of our framework, namely sampling, keyframe management, and training.

\noindent\textbf{Training }is continuously conducted in real-time to optimize objectives that include common photometric and geometric losses in accordance with~\cref{eq:nerf_loss}, as well as SDF loss and free space suppressing loss that are used in previous SDF-based INR-SLAM~\cite{coslam23,eslam23}. Detailed formulations are available in the \textit{supplmat}.

\noindent\textbf{Sampling.}
For pixels with available depth, we utilize stratified sampling, dividing the samples into two categories: surface and free space. Surface samples are densely placed around the ground truth depth to accurately capture the surface details within the truncation range. Free space samples, on the other hand, are evenly distributed along the ray. In cases where ground truth depth is absent, we evenly allocate samples.
This method highlights the fine details of surfaces, making the use of an \( L2 \) loss more effective than the \( L1 \) loss traditionally used in earlier studies \cite{imap22,voxfusion22,niceslam22}.
%

\noindent\textbf{Keyframe Selection.}
Enhanced pose estimation and consistent global reconstruction can be achieved with a global bundle adjustment strategy, as suggested in~\cite{goslam23,coslam23}. For better storage and retrieval efficiency, we follow the approach in~\cite{coslam23}, selectively caching only key sampled rays from keyframes for optimization purposes.

\subsection{Evaluation Metrics}
\label{subsec:metric}

For overall performance, We evaluate the final reconstruction quality using three established metrics that are commonly used in INR-based RGB-D SLAM~\cite{imap22,niceslam22,coslam23,eslam23}: Accuracy(\textit{Acc.}[cm]), Completion(\textit{Comp.}[cm]), and Completion Ratio that gauges the proportion of extracted meshes that of Completion value smaller than 5cm (\textit{Comp.}[\%]). For the sampled points on the reconstructed mesh and the ground truth mesh, the Accuracy and Completion metrics are calculated by determining the average distance from the former to the latter, and from the latter to the former, respectively.
Prior to metric calculation, mesh culling is conducted in line with the procedure outlined in~\cite{neuralrgbd22}. 
The overall accuracy of the camera trajectory is quantified by the Root Mean Square Error (\textit{ATE}[cm] \textit{RMSE}). 

For procedural performance, we assess the Peak Signal-to-Noise Ratio (\textit{PSNR}[db]) and L1 term of the estimated depth (\textit{Depth L1}[cm]) throughout the entire sequence by comparing their mean values, since SLAM additionally emphasizes the continual estimation performance, unlike static 3D reconstruction tasks.
The efficiency of mapping and tracking is evaluated by the average update time per input frame over the whole sequence.

\subsection{Scenario Settings}
\label{sub:scenario}
As illustrated in~\cref{fig:main_framework}, our goal is to incrementally unveil the capabilities of INRs for RGB-D SLAM by implementing a unified evaluation protocol. Thus, it is possible to identify the best choices for subsequent refinements. To accommodate the extensive range of evaluation metrics and inherent complexity of SLAM systems, we establish two distinct scenarios, each with a specific evaluative emphasis, enhancing the modularity and clarity of our analysis:

\noindent\textbf{Lab Scenario.}
This scenario aims to gain comprehensive insights from a detailed leaderboard, see Tab.~\ref{tab:main_leaderboard}, based on the synthetic Replica dataset~\cite{replica19arxiv}. The initial evaluation centers on the structural paradigm of INRs, with the best-performed one being adopted for subsequent benchmark analysis. Then, we construct the main leaderboard using eight synthetic sequences from the Replica dataset. Each sequence includes ground truth trajectories, guaranteeing thorough coverage of the entire room.

\noindent\textbf{Practical Scenario.}
In contrast to simulated datasets, real-world data are subject to noise. Also, camera trajectories often only capture the scenes partially, posing distinct challenges for INR-based RGB-D SLAM systems.
For this practical scenario, we evaluate seven sequences from the synthetic NeuralRGBD dataset~\cite{neuralrgbd22}, which are crafted to mimic the noise and artifacts characteristic of real-world depth sensors. Intentionally, the camera trajectories within these experiments were designed to scan only portions of the scenes. The leaderboard results are shown in Tab.~\ref{tab:synthetic_leaderboard}.
Notably, the metrics of completeness and accuracy typically used for full-scene reconstruction are less definitive in the context of partial observations. Nonetheless, prior research suggests that increased completeness correlates with improved pose estimation. Therefore, in this scenario, the reconstruction quality (\textit{Acc.} and \textit{Comp.}) 
act only as indicators for the SLAM performance. We instead place significant emphasis on performance metrics such as \textit{ATE}, \textit{Depth L1}, and \textit{PSNR}.

Note that, due to substantial variations in sequence lengths, it introduces considerable variability in average frame processing time. Therefore, we remove the tracking and mapping speed metrics from the practical scenario leaderboard for more transparent comparisons. The full records of time measurement can be found in the \textit{supplmat}.

%
\subsection{Optimal INR Designs}
\label{sub:new_design}
We concentrate on the \textbf{NeRF-centric} paradigm to showcase two distinct scenarios in our leaderboard. It provides critical insights for developing an optimal INR. \textit{We will introduce our new design in conjunction with the discussion in the experimental section}.

From the \textbf{SLAM-centric} viewpoint, NeRF shows promise as a dual-purpose tool for localization and mapping. However, its effectiveness is somewhat limited in applications, such as robot navigation. 
This domain necessitates a SLAM system that is not only \textit{robust} but also capable of \textit{rapid convergence}, particularly in environments with ambiguously defined scene boundaries. To address these limitations, we propose to \textit{integrate the insights from the NeRF-centric leaderboard into SLAM-centric paradigms}. We provide a qualitative evaluation (see~\cref{fig:real}) on Scannet Dataset~\cite{scannet17}, demonstrating its adaptability and enhanced performance within the context of SLAM-centric methods.

\tabcolsep=0.1cm
\begin{table*}[t!]
\centering
\small
\scalebox{0.9}{
\begin{tabular}{cccccccccc}
\toprule
\multirow{2}{*}{$\mathcal{G}$} & \multirow{2}{*}{$\mathcal{F}$} & \multicolumn{4}{c}{From results} & \multicolumn{4}{c}{From processes}\\
\cmidrule(lr){3-6}
\cmidrule(lr){7-10}
 & & Acc.{[}cm{]}$\downarrow$ & Comp.{[}cm{]}$\downarrow$ & Comp.[$\%$]$\uparrow$ & ATE{[}cm{]}$\downarrow$ & PSNR{[}db{]}$\uparrow$ & Depth L1{[}cm{]}$\downarrow$ & Tracking{[}ms{]}$\downarrow$ & Mapping{[}ms{]}$\downarrow$ \\ 
\hline
\multirow{5}{*}{\rotatebox{90}{SDF(direct)}}  
                            & MLP & 11.95& 8.36& 76.82& 14.41& 24.20& 3.12& 293&421 \\
                            & \textbf{Dense$^1$} &  \color{blue}{\textbf{1.65}}& 5.62&  \color{blue}{\textbf{83.93}}&  \color{blue}{\textbf{1.37}} & 27.88& \textbf{1.50}& 288& \textbf{286}\\
                            & Sparse$^3$ & 1.76& 5.66& 83.61& 1.40                 & \textbf{28.23}&  \color{blue}{\textbf{1.65}}& \textbf{197}&  \color{blue}{\textbf{300}}\\
                            & Tri. & 1.69& 5.64& 83.60& 1.42                   & 27.52& 1.80& 352& 820\\
                            & Fact. & 1.69& 5.60& 83.69& 1.50                  & 27.52& 1.74& 419& 911\\
\hline
\multirow{5}{*}{\rotatebox{90}{SDF(density)}}  
                            & MLP & 9.64& 9.92& 72.22& 24.58& 23.51& 7.01& 250& 419\\
                            & \textbf{Dense$^2$} & \textbf{1.60}& \textbf{5.58}& \textbf{84.01}& \textbf{1.31}                & 27.77& 4.42& 253& 585\\
                            & Sparse$^4$ & 1.69& 5.65& 83.72& 1.40 & \color{blue}{\textbf{28.10}}& 4.45&  \color{blue}{\textbf{207}}& 310\\
                            & Tri.$^5$ & 1.80&  \color{blue}{\textbf{5.59}}& 83.82& 1.49                   & 27.55& 4.51& 376& 829\\
                            & Fact. & 1.75& 5.60& 83.73& 1.55                  & 27.54& 4.48& 414& 897\\
\hline
\multirow{5}{*}{\rotatebox{90}{SDF(surface)}}  
                            & MLP & 30.44& 24.28& 20.21& 44.68& 17.13& 98.24& 342& 585\\
                            & Dense & 32.83& 20.71& 42.28& 135.39             & 16.21& 176.15& 418& 669\\
                            & Sparse & 48.22& 25.73& 30.73& 176.68            & 16.49& 174.94& 258& 359\\
                            & Tri. & 30.28& 18.06& 41.77& 87.71               & 16.05& 180.68& 490& 897\\
                            & Fact. & 30.75& 20.45& 31.31& 85.63              & 16.43& 168.81& 584& 1009\\
                            
\bottomrule
\end{tabular}
}
\vspace{-8pt}
\captionsetup{justification=justified}
\caption{\textbf{The leaderboard of lab scenario}. Text in \textbf{bold} indicates the best performance, while text in \textbf{\textcolor{blue}{blue bold}} denotes the second best. $^{1-5}$ indicated the top 5 combinations according to the counting of performance ranking.}
\label{tab:main_leaderboard}
\vspace{-10pt}
\end{table*}
\vspace{-2pt}
\section{Experiments and New Designs}
\vspace{-2pt}

In this section, we firstly observe the impact of various combinations of NeRF components on SLAM system performance under a Lab scenario. Subsequently, we examine whether the system's performance alters in a practical scenario. Finally, based on the analysis of these empirical observations, which are displayed as leaderboards in~\cref{sub:lead}, we propose new NeRF designs suitable for different SLAM scenarios in~\cref{sub:exp_new}.
Please refer to the \textit{supplmat} for implementation specifications, including parameters and platform configurations. Note that 'Tri-plane' and 'Factorization' are denoted as Tri. and Fact. in this section, respectively.

\subsection{Benchmark Leaderboard}
\label{sub:lead}
\textbf{Lab Scenario.}
We begin by examining the impact of structural paradigms, which are commonly presumed but not often analyzed within the current NeRF-SLAM paradigm. In line with~\cref{eq:nerf_state}, we list the results in~\cref{tab:coupling_F}, focusing on the \textit{initialization phase}, where INRs are trained using a single posed RGB-D frame.
\tabcolsep=0.045cm
\begin{table}[t!]
\small
\centering
\scalebox{0.85}{
\begin{tabular}{cccccc}
\toprule
$\mathcal{F}$   & Structure  & PSNR[db] $\uparrow$ & Depth L1[cm] $\downarrow$  & Time[s] $\downarrow$ \\ 
\hline
\multirow{3}{*}{MLP}    &Coupled(base)          &  23.03            &2.19           & \textbf{65.59} \\
                        &\textbf{Coupled}       &  \textbf{23.76}   &\textbf{1.68}  & 82.00  \\
                        &Decoupled              &  6.37             &2.26           & 106.7   \\
\hline
\multirow{3}{*}{Dense}  &\textbf{Coupled(base)} &  30.21            & \textbf{0.83}     & \textbf{47.55} \\
                        &Coupled                &  29.66            &0.84               &  93.00 \\
                        &Decoupled              &  \textbf{31.12}   &0.85            & 85.76 \\
\hline
\multirow{3}{*}{Sparse} &Coupled(base)          &  28.28            &0.97               & 49.31 \\
                        &Coupled                &  28.45            &\textbf{0.92}      & 70.27 \\
                        &\textbf{Decoupled}     &  \textbf{30.42}   &0.99             & \textbf{32.72} \\
\hline
\multirow{3}{*}{Tri.}   &Coupled(base)          & 26.41             &1.17           & \textbf{61.78} \\
                        &Coupled                &  25.45            &1.16           & 120.40 \\
                        &\textbf{Decoupled}     &  \textbf{27.95}   &\textbf{1.08}  & 83.22  \\
\hline
\multirow{3}{*}{Fact.}  &Coupled(base)          &  26.34            &1.19           & 121.10 \\
                        &Coupled                &  25.58            &1.28           & 127.40 \\
                        &\textbf{Decoupled}     &\textbf{28.36}     &\textbf{1.07}  & \textbf{85.83}  \\
\bottomrule
\end{tabular}
}
\vspace{-8pt}
\captionsetup{justification=justified}
\caption{\textbf{Impact of network architectures at initialization}: performance comparison on Replica Room0 sequence, where Time[s] indicates the total processing time in seconds. All outcomes correspond to the geometric function SDF (direct).}
\vspace{-12pt}
\label{tab:coupling_F}
\end{table}
%
It can be seen that coupling geometric and appearance features boosts depth estimation within a limited number of iterations for MLP, dense, and sparse representations, suggesting a faster 
SDF convergence. MLP notably shows about a 25\% reduction in \textit{Depth L1} loss when adopting a coupled structure. However, the coupled paradigm seems to compromise color rendering in dense (0.91dB \textcolor{red}{$\downarrow$}) and Sparse (1.97dB \textcolor{red}{$\downarrow$}) methods. On the contrary, decomposition methods (Tri. and Fact.) benefit from decoupled structures, showing both enhanced geometric (0.08cm and 0.12cm \textcolor{red}{$\downarrow$} in \textit{Depth L1} loss) and appearance (1.54dB and 2.02dB \textcolor{red}{$\uparrow$}) accuracy compared to their coupled counterparts.


The structure with the most top-ranked instances (highlighted in \textbf{bold}) in~\cref{tab:coupling_F} is selected as the optimal choice for each INR (\(\mathcal{F}\)) and is used to construct the following leaderboards. We assess the collective performance of \(\mathcal{F}\) and \(\mathcal{G}\) within the unified SLAM framework, with results for the Lab scenario detailed in~\cref{tab:main_leaderboard}.

Overall, hybrid representations (\ie, $\mathcal{F}$ other than MLP) demonstrate markedly superior color and geometry estimation performance, with decomposition methods (Tri. and Fact.) generally lagging behind grid-based ones (dense and sparse) in terms of processing speed, shown in both~\cref{tab:coupling_F} and~\cref{tab:main_leaderboard}. 
Despite using a coupled network structure that halves the total optimization parameters, \textit{the dense grid representation excels}. It achieves overall six top and three second-place rankings. This is \textit{followed by sparse encoding}, which secures two top spots and four second-place, with the Tri.  claiming one second-place spot.
It is worth noting that the dense grid's concatenated 4D features consistently yield excellent SDF results (with first and second rankings in both \textit{Acc.} and \textit{Depth L1}) and hold their own in RGB quality (marginally lower than the 8D feature-encoded sparse grid) across the SLAM process while maintaining the quickest mapping speed.

Notably, the neural implicit surfaces~\cite{neus21} rendering method shows significantly inferior performance than its more naive counterparts~\cite{neuralrgbd22}, which also shares the unbiased approximation but directly maps SDF to weighting factors rather through volume rendering. This finding might suggest the delicate formulation for occlusion in offline 3D volume rendering might be sensitive to relatively poorly estimated poses in the online SLAM.


\noindent\textbf{Practical Scenario.}
\tabcolsep=0.045cm
\begin{table}[t!]
\small
\centering
\setlength{\tabcolsep}{0.5mm}{
\scalebox{0.9}{
\begin{tabular}{ccccc|ccc}
\toprule
 &  & \multicolumn{3}{c}{Indicators} & \multicolumn{3}{c}{Targets}\\
\cmidrule(lr){3-5}
\cmidrule(lr){6-8}
$\mathcal{G}$
& $\mathcal{F}$ 
& \makecell[c]{ Acc.$\downarrow$ \\{[}cm{]}}
& \makecell[c]{ Comp.$\downarrow$ \\{[}cm{]}}
& \makecell[c]{ Comp.$\uparrow$ \\{[}$\%${]}}
& \makecell[c]{ ATE$\downarrow$ \\{[}cm{]}}
& \makecell[c]{ PSNR$\uparrow$ \\{[}db{]}}
& \makecell[c]{ Depth L1$\downarrow$ \\{[}cm{]}}\\
\hline
\multirow{6}{*}{\rotatebox{90}{SDF(direct)}}  
                            & MLP & 4.37& 5.16& 79.22& 3.69 &22.56 &3.56\\
                            & Dense & 2.69& 4.69& 83.45& 1.96 &\underline{25.15} &\color{blue}{\textbf{1.70}}\\
                            & Sparse & 2.84& 4.81& 82.64& 2.12 & \color{blue}{\textbf{25.23}} &\underline{1.84}\\
                            & Tri. & 2.29& \textbf{4.42}& \textbf{84.01}& \underline{1.89} &24.67 &1.85\\
                            & Fact. & 2.62& \color{blue}{\textbf{4.47}}& 83.54& 1.94 &24.69 &1.87\\
                            & \textbf{Hybrid} & 2.40& 4.64& 83.48& \textbf{1.86}& \textbf{25.25}& \textbf{1.68}\\
\hline
\multirow{6}{*}{\rotatebox{90}{SDF(density)}}  
                            & MLP & 3.98& 5.12& 78.82& 3.87 &22.56 &4.80\\
                            & Dense & 2.70& 4.72& 83.27&\color{blue}{\textbf{1.87}} &25.04 &4.40\\
                            & Sparse & 2.84& 4.71& 82.79& 1.96 &25.08 &4.46\\
                            & Tri. & \color{blue}{\textbf{2.12}}& 4.62& \color{blue}{\textbf{83.90}}& 1.90 &24.62 &4.42\\
                            & Fact. & \textbf{2.11}& 4.45& \textbf{84.01}& 2.01 &24.63 &4.41\\
                            & Hybrid & 2.34& 4.71& 83.25& 1.91& 25.05&4.40 \\
\bottomrule
\end{tabular}
}}
\vspace{-8pt}
\captionsetup{justification=justified}
\caption{\textbf{The leaderboard of practical scenario}. Black bold text indicates top performance, blue marks second place, with an additional underline denotes third rank for the targeted metric. Note that the hybrid denotes our proposed encoding strategy that synergizes the strengths of dense grid and tri-plane.}
\vspace{-10pt}
\label{tab:synthetic_leaderboard}
\end{table}
%
When adopting synthetic data with noisy depth and incomplete scene coverage, decomposition methods excel in achieving high geometric accuracy (\textit{Acc.}) and scene completion (\textit{Comp.})~\cref{tab:synthetic_leaderboard}. 
This verifies our statement in~\cref{sub:scenario}. The decomposition of 3D implicit spaces into mutually orthogonal 2D (additional 1D for the factorization method) results in planar geometries akin to real-world indoor environments, extending even beyond the observed view frustum. In these instances, more points might align closely with the true spatial geometry.

Yet, this increase in scene completion doesn't directly improve targeting performance metrics. Specifically, \textit{dense representation still performs better} in position accuracy (\textit{ATE}) and depth estimation (\textit{Depth L1}), compared to the decomposition methods, especially when combined with SDF(direct).
In the realm of color estimation, dense grid exhibit only a marginal drop in accuracy (0.08dB\textcolor{red}{$\downarrow$}) compared to their sparse counterparts. This trend is consistent with the Lab scenario leaderboard~\cref{tab:coupling_F}, where both grid-based methods notably surpass the performance of decomposition methods in \textit{ATE}, \textit{Depth L1} and \textit{PSNR}.



\subsection{New Designs for Encoding}
\label{sub:exp_new}

\tabcolsep=0.065cm
\begin{figure}[t!]
  \centering
   \includegraphics[width=1\linewidth]{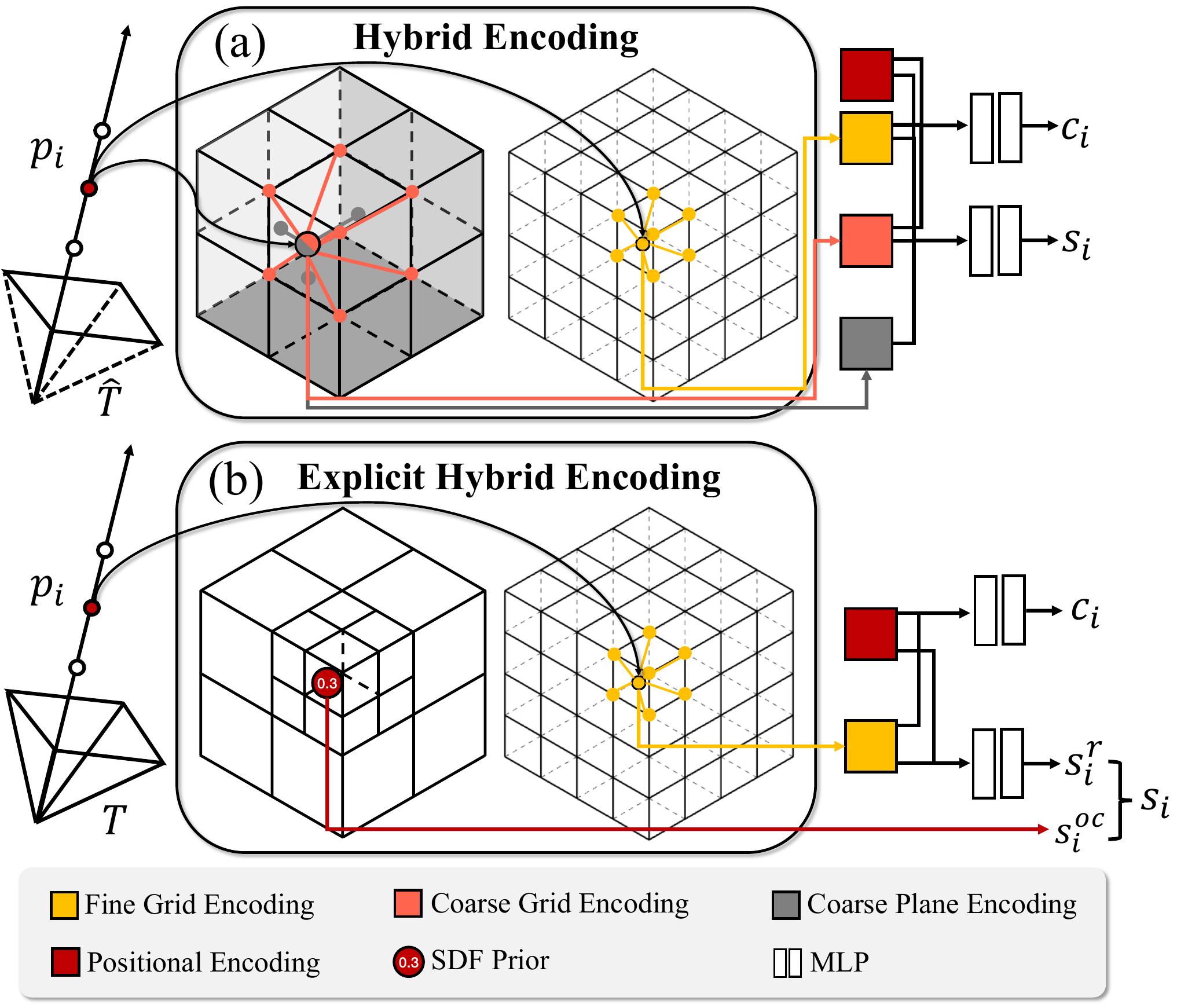}
   \vspace{-17pt}
   \caption{\textbf{Illustration of new designs.} For hybrid encoding, a point \( p_i \) is (a) encoded using feature planes and a feature grid at a coarse level, and exclusively by a feature grid at a fine level. In contrast, for explicit hybrid encoding, \( p_i \) is (b) solely encoded with an optimizable fine-level feature grid and decoded by MLP into an SDF residual \( s^r_i \) and color \( c_i \). This residual is then combined with the SDF prior stored in an explicit octree \( s^{oc}_i \) to derive the inferred SDF value \( s_i \).}
   \vspace{-10pt}
   \label{fig:hybrid_diagram}
\end{figure}
\tabcolsep=0.06cm
\begin{figure}[t!]
  \centering
   \includegraphics[width=1\linewidth]{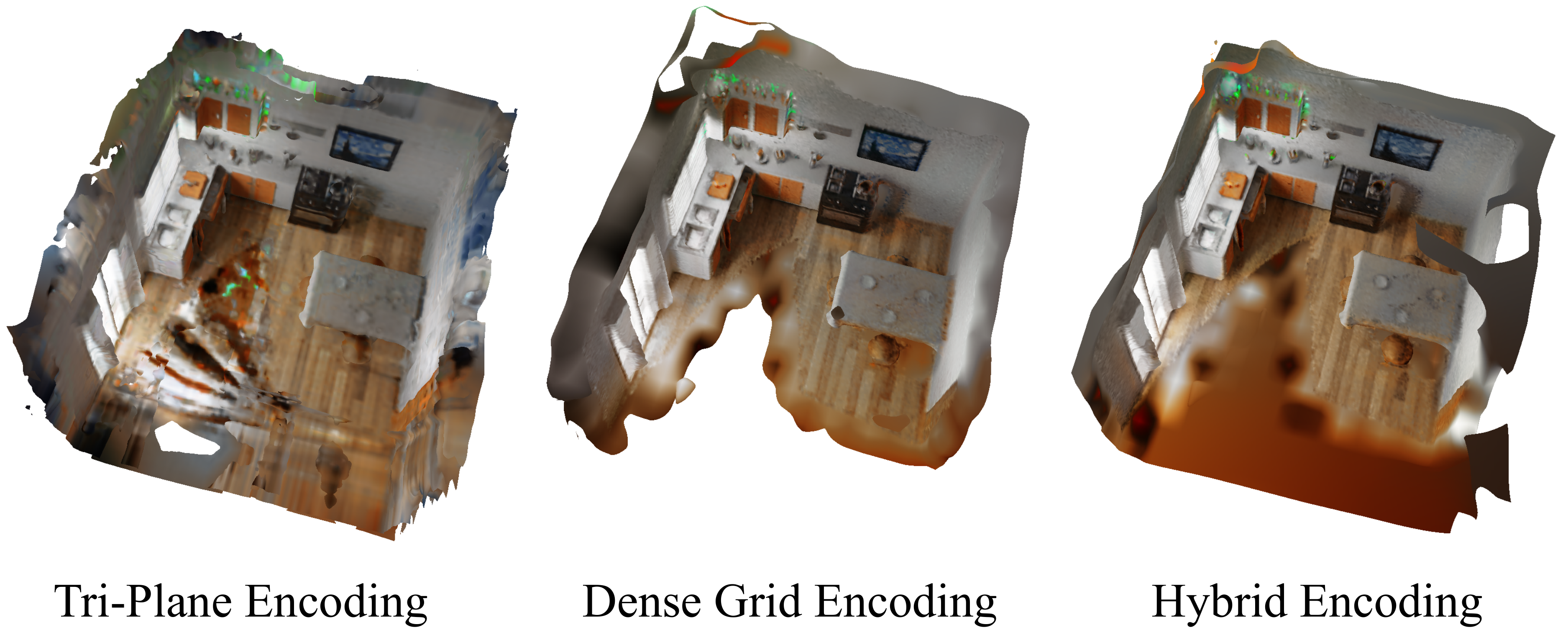}
   \vspace{-18pt}
   \caption{Reconstruction of `\textit{morning apartment}' sequence on the NeuralRGBD dataset, Our hybrid encoding strategy brings the best of two worlds.}
   \vspace{-10pt}
   \label{fig:synthetic_ma}
\end{figure}
\begin{figure*}[t!]
  \centering
   \includegraphics[width=1\linewidth]{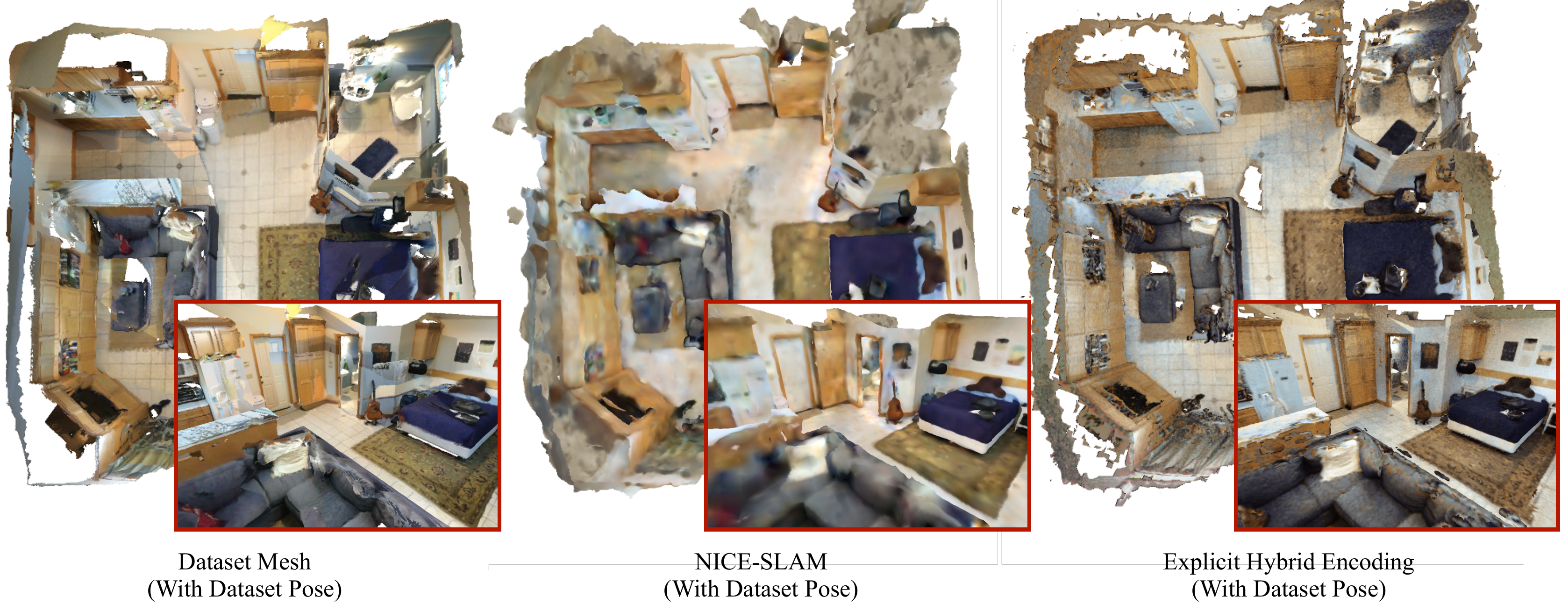}
   \vspace{-20pt}
   \caption{Qualitative evaluation of explicit hybrid encoding on `\textit{scene0000}' sequence of ScanNet Dataset. Both NICE-SLAM and Ours run on the posed RGB-D stream to simulate an externally provided tracker.}
   \vspace{-14pt}
   \label{fig:real}
\end{figure*}

\noindent \textbf{Hybrid Encoding.} NeRF-centric approaches estimate camera trajectories by freezing the parameters of $\mathcal{F}$ and setting 6-DOF poses as optimizable parameters. Therefore, effective modelling of spatial geometry and color is beneficial for pose estimation. 
Decomposition-based methods encode scenes in ways that echo the real three-dimensional world, particularly within artificial indoor settings. This characteristic strengthens their capacity to project unobserved regions accurately, allowing for sturdier pose estimates even when faced with occasional inaccuracies in observed segments of the scene. However, their precision falls short of grid-based methods, leading us to consider combining the strengths of both approaches.
For this reason, we introduce \textit{hybrid encoding, a strategy blending the comprehensive spatial depiction of decomposition with the precision of grid-based methods}, as depicted in~\cref{fig:hybrid_diagram} (a). This strategy complies with our leaderboard protocol, with both feature dimensions and resolution levels set at two. 
At the coarse resolution level, a sample point undergoes both bilinear and trilinear interpolation within the respective feature plane and grid. At the fine resolution level, interpolation is exclusively performed using a feature grid. Outputs from both scales are then combined to form the input for MLP decoders for estimating raw color and geometry.

The quantitative and qualitative results are shown in ~\cref{tab:synthetic_leaderboard} and~\cref{fig:synthetic_ma} for the dense grid feature encoding. It is revealed that hybrid encoding achieves superior trajectory estimation accuracy and reconstruction fidelity, both in color and depth. This is mainly due to the enhanced completeness augmented by plane-based representation, as confirmed by various indicators, \ie, increased performance for \textit{Acc.} and \textit{Comp.}, and visual demonstration in~\cref{fig:synthetic_ma}.

Notably, our benchmark strictly controls feature dimensions and spatial resolution for the implicit encoding. However, optimal performance in such a controlled comparisons may not always align with real-world application priorities. For example, a minor compromise in accuracy (\eg a 1mm decrease in trajectory estimation accuracy) can lead to substantial gains in memory efficiency. Hence, \textit{hybrid encoding combining tri-plane and hash grid feature encoding} might be a preferable alternative to the tri-plane and dense combination. Such trade-off is discussed in the \textit{supplmat}.

\noindent \textbf{Explicit Hybrid Encoding.}
SLAM-centric methods utilize external trackers to ensure robust pose estimation and maintain system stability over time. Our evaluations show that dense grids, along with their hybrid encoding with tri-plane, excel in implicit scene encoding across two prominent leaderboards. To evolve these techniques for NeRF-based mapping, enhancements in computational and memory efficiency are necessary. Dense grids exhibit cubic scaling with resolution (\(O(n^3)\)), while octrees show logarithmic scaling (\(O(log(n))\)) in sparse environments, leaving room for memory optimization. 
Drawing inspiration from ~\cite{voxfusion22,shine-map23}, we propose an explicit hybrid encoding, see ~\cref{fig:hybrid_diagram}(b), to \textit{substitute the coarse-level feature grid with an octree structure and simplifying the encoding process by using a single-level dense grid}. To reduce the complexity of simultaneously encoding color and geometry for real-time applications using only two-dimensional features, we adopt the SDF residual optimization strategy detailed in~\cite{H2Mapping23}.
%

Our explicit hybrid encoding method's capability for high-fidelity online map updating is qualitatively showcased in~\cref{fig:real}. When tested on sequence \textit{'scene0169'} with a posed RGB-D stream on our hardware platform, explicit hybrid encoding achieves map updating at approximately 1 Hz, compared to around 0.25 Hz for NICE-SLAM. More records are available in \textit{supplmat}.

\tabcolsep=0.045cm
\begin{table}[t!]
\small
\centering
\scalebox{0.85}{
\begin{tabular}{ccc|cc}
\toprule
 &  \multicolumn{2}{c|}{Lab} & \multicolumn{2}{c}{Practical}\\
\cmidrule(lr){2-3}
\cmidrule(lr){4-5}
&NICE-SLAM~\cite{niceslam22}
& \makecell[c]{Ours}
&CO-SLAM~\cite{coslam23}
& \makecell[c]{Ours}\\
\hline
 Depth L1$\downarrow$   & 3.53 & \textbf{1.50}& 3.02& \textbf{1.68}\\
 Acc.$\downarrow$       & 2.85 & \textbf{1.65}& 2.95& \textbf{2.40}\\
 Comp.$\downarrow$      & \textbf{3.00} & 5.62& \textbf{2.96}& 4.64\\
 Comp.\%$\uparrow$      & \textbf{89.33}& 83.93& \textbf{86.88}& 83.48\\
 ATE$\downarrow$        & 1.95& \textbf{1.37}& -& \textbf{1.86}\\
 Res$\times$Dim$\downarrow$ & 3$\times$32& \textbf{2$\times$2}& 16$\times$2& \textbf{2$\times$2}\\
\bottomrule
\end{tabular}
}
\vspace{-8pt}
\captionsetup{justification=justified}
\caption{
Quantitative comparison with the SOTA method, Co-SLAM, featuring similar implicit scene representations. In the Lab scenario, both NICE-SLAM and our approach use dense grid representations, whereas, in the practical scenario, Co-SLAM and Our method are both in the spirit of hybrid encoding.}
\vspace{-14pt}
\label{tab:sota}
\end{table}


\noindent\textbf{Discussion.}
For context, we include comparisons with existing methods in~\cref{tab:sota}. The top performances, \ie, dense grid encoding with SDF(density) rendering in the Lab scenarios, and hybrid encoding with SDF(direct) rendering in the Practical scenarios showcase higher reconstruction and pose estimation accuracy, even with generally lower completeness and notably fewer total feature dimensions (Res$\times$Dim). For a visual representation of these findings, readers are directed to the \textit{supplmat}.
%
In this benchmark, discussions concentrate on achieving optimal performance, rigorously measured against specific metrics under strictly regulated variables for fairness comparison. Nonetheless, the appeal of computational efficiency often takes precedence over the pursuit of complete comparative fairness. Please refer to the \textit{supplmat} for such trade-offs.

\vspace{-3pt}
\section{Conclusion and Future Work}
\vspace{-3pt}

We proposed an open-source benchmark to evaluate INRs for RGB-D SLAM, filling a crucial gap in standardizing performance assessments. We demonstrated the superior efficacy of dense grid representations and introduced a hybrid encoding strategy that marries precision with efficiency. Our work not only guided the selection of INR components but also advanced practical SLAM applications.

\noindent\textbf{Future Work.} 
SLAM systems are intricate and are often tailored to specific environments. Consequently, it is not feasible to declare a universally superior representation within the scope of this paper. For example, in scenarios with orthogonal geometry and muted colors, such as an office corridor, planar representations may stand as more suitable alternatives to the dense grids. We advise that future research should expand to more diverse scenes.

\noindent \textbf{Acknowledgement.} This paper is supported by the National Natural Science Foundation of China (NSF) under Grant No. NSFC22FYT45 and the Guangzhou City, University and Enterprise Joint Fund under Grant No.SL2022A03J01278.

\clearpage
{
    \small
    \bibliographystyle{ieeenat_fullname}
    \bibliography{main}

\begin{thebibliography}{59}
\providecommand{\natexlab}[1]{#1}
\providecommand{\url}[1]{\texttt{#1}}
\expandafter\ifx\csname urlstyle\endcsname\relax
  \providecommand{\doi}[1]{doi: #1}\else
  \providecommand{\doi}{doi: \begingroup \urlstyle{rm}\Url}\fi

\bibitem[Azinovi{\'c} et~al.(2022)Azinovi{\'c}, Martin-Brualla, Goldman, Nie{\ss}ner, and Thies]{neuralrgbd22}
Dejan Azinovi{\'c}, Ricardo Martin-Brualla, Dan~B Goldman, Matthias Nie{\ss}ner, and Justus Thies.
\newblock Neural rgb-d surface reconstruction.
\newblock In \emph{Proceedings of the IEEE/CVF Conference on Computer Vision and Pattern Recognition}, pages 6290--6301, 2022.

\bibitem[Bian et~al.(2023)Bian, Wang, Li, Bian, and Prisacariu]{pose3}
Wenjing Bian, Zirui Wang, Kejie Li, Jia-Wang Bian, and Victor~Adrian Prisacariu.
\newblock Nope-nerf: Optimising neural radiance field with no pose prior.
\newblock In \emph{Proceedings of the IEEE/CVF Conference on Computer Vision and Pattern Recognition}, pages 4160--4169, 2023.

\bibitem[Campos et~al.(2021)Campos, Elvira, Rodr{\'\i}guez, Montiel, and Tard{\'o}s]{tra_21orb3}
Carlos Campos, Richard Elvira, Juan J~G{\'o}mez Rodr{\'\i}guez, Jos{\'e}~MM Montiel, and Juan~D Tard{\'o}s.
\newblock Orb-slam3: An accurate open-source library for visual, visual--inertial, and multimap slam.
\newblock \emph{IEEE Transactions on Robotics}, 37\penalty0 (6):\penalty0 1874--1890, 2021.

\bibitem[Chan et~al.(2022)Chan, Lin, Chan, Nagano, Pan, De~Mello, Gallo, Guibas, Tremblay, Khamis, et~al.]{triplane21}
Eric~R Chan, Connor~Z Lin, Matthew~A Chan, Koki Nagano, Boxiao Pan, Shalini De~Mello, Orazio Gallo, Leonidas~J Guibas, Jonathan Tremblay, Sameh Khamis, et~al.
\newblock Efficient geometry-aware 3d generative adversarial networks.
\newblock In \emph{Proceedings of the IEEE/CVF Conference on Computer Vision and Pattern Recognition}, pages 16123--16133, 2022.

\bibitem[Chen and Lee(2023)]{pose4}
Yu Chen and Gim~Hee Lee.
\newblock Dbarf: Deep bundle-adjusting generalizable neural radiance fields.
\newblock In \emph{Proceedings of the IEEE/CVF Conference on Computer Vision and Pattern Recognition}, pages 24--34, 2023.

\bibitem[Chung et~al.(2023)Chung, Tseng, Hsu, Shi, Hua, Yeh, Chen, Chen, and Hsu]{orbeez23}
Chi-Ming Chung, Yang-Che Tseng, Ya-Ching Hsu, Xiang-Qian Shi, Yun-Hung Hua, Jia-Fong Yeh, Wen-Chin Chen, Yi-Ting Chen, and Winston~H Hsu.
\newblock Orbeez-slam: A real-time monocular visual slam with orb features and nerf-realized mapping.
\newblock In \emph{2023 IEEE International Conference on Robotics and Automation (ICRA)}, pages 9400--9406. IEEE, 2023.

\bibitem[Cole et~al.(2021)Cole, Genova, Sud, Vlasic, and Zhang]{cole2021differentiable}
Forrester Cole, Kyle Genova, Avneesh Sud, Daniel Vlasic, and Zhoutong Zhang.
\newblock Differentiable surface rendering via non-differentiable sampling.
\newblock In \emph{Proceedings of the IEEE/CVF International Conference on Computer Vision}, pages 6088--6097, 2021.

\bibitem[Dai et~al.(2016)Dai, Nie{\ss}ner, Zollh{\"o}fer, Izadi, and Theobalt]{RGBDSLAM6}
Angela Dai, Matthias Nie{\ss}ner, Michael Zollh{\"o}fer, Shahram Izadi, and Christian Theobalt.
\newblock Bundlefusion.
\newblock \emph{ACM Transactions on Graphics (TOG)}, 36:\penalty0 1 -- 18, 2016.

\bibitem[Dai et~al.(2017)Dai, Chang, Savva, Halber, Funkhouser, and Nie{\ss}ner]{scannet17}
Angela Dai, Angel~X Chang, Manolis Savva, Maciej Halber, Thomas Funkhouser, and Matthias Nie{\ss}ner.
\newblock Scannet: Richly-annotated 3d reconstructions of indoor scenes.
\newblock In \emph{Proceedings of the IEEE conference on computer vision and pattern recognition}, pages 5828--5839, 2017.

\bibitem[Deng et~al.(2023)Deng, Wu, Chen, Xia, Sun, Liu, Yu, and Pei]{nerf-loam23}
Junyuan Deng, Qi Wu, Xieyuanli Chen, Songpengcheng Xia, Zhen Sun, Guoqing Liu, Wenxian Yu, and Ling Pei.
\newblock Nerf-loam: Neural implicit representation for large-scale incremental lidar odometry and mapping.
\newblock In \emph{Proceedings of the IEEE/CVF International Conference on Computer Vision}, pages 8218--8227, 2023.

\bibitem[Engel et~al.(2014)Engel, Sch{\"o}ps, and Cremers]{tra_Engel2014LSDSLAMLD}
Jakob~J. Engel, Thomas Sch{\"o}ps, and Daniel Cremers.
\newblock Lsd-slam: Large-scale direct monocular slam.
\newblock In \emph{European Conference on Computer Vision}, 2014.

\bibitem[Haghighi et~al.(2023)Haghighi, Kumar, Thiran, and Van~Gool]{task_semantic23}
Yasaman Haghighi, Suryansh Kumar, Jean~Philippe Thiran, and Luc Van~Gool.
\newblock Neural implicit dense semantic slam.
\newblock \emph{arXiv preprint arXiv:2304.14560}, 2023.

\bibitem[Hu et~al.(2023)Hu, Mao, Bao, Zhang, and Cui]{cpslam23}
Jiarui Hu, Mao Mao, Hujun Bao, Guofeng Zhang, and Zhaopeng Cui.
\newblock Cp-slam: Collaborative neural point-based slam system.
\newblock In \emph{Thirty-seventh Conference on Neural Information Processing Systems}, 2023.

\bibitem[Hua et~al.(2023)Hua, Bai, Cao, and Wang]{fmapping23}
Tongyan Hua, Haotian Bai, Zidong Cao, and Lin Wang.
\newblock Fmapping: Factorized efficient neural field mapping for real-time dense rgb slam.
\newblock \emph{arXiv preprint arXiv:2306.00579}, 2023.

\bibitem[Hua et~al.(2024)Hua, Bai, Cao, Liu, Tao, and Wang]{hua2024himap}
Tongyan Hua, Haotian Bai, Zidong Cao, Ming Liu, Dacheng Tao, and Lin Wang.
\newblock Hi-map: Hierarchical factorized radiance field for high-fidelity monocular dense mapping.
\newblock \emph{arXiv preprint arXiv:2401.03203}, 2024.

\bibitem[Jeong et~al.(2021)Jeong, Ahn, Choy, Anandkumar, Cho, and Park]{pose1}
Yoonwoo Jeong, Seokjun Ahn, Christopher Choy, Anima Anandkumar, Minsu Cho, and Jaesik Park.
\newblock Self-calibrating neural radiance fields.
\newblock In \emph{Proceedings of the IEEE/CVF International Conference on Computer Vision}, pages 5846--5854, 2021.

\bibitem[Jiang et~al.(2023)Jiang, Zhang, Liu, Yu, Cheng, Zhou, and Shen]{H2Mapping23}
Chenxing Jiang, Han-Qi Zhang, Peize Liu, Zehuan Yu, Hui Cheng, Boyu Zhou, and Shaojie Shen.
\newblock H\$\_\{2\}\$-mapping: Real-time dense mapping using hierarchical hybrid representation.
\newblock \emph{IEEE Robotics and Automation Letters}, 8:\penalty0 6787--6794, 2023.

\bibitem[Johari et~al.(2022)Johari, Carta, and Fleuret]{eslam23}
Mohammad~Mahdi Johari, Camilla Carta, and Fran{\c{c}}ois Fleuret.
\newblock {ESLAM:} efficient dense {SLAM} system based on hybrid representation of signed distance fields.
\newblock \emph{CoRR}, abs/2211.11704, 2022.

\bibitem[Kerbl et~al.(2023)Kerbl, Kopanas, Leimk{\"u}hler, and Drettakis]{3dgaussian23}
Bernhard Kerbl, Georgios Kopanas, Thomas Leimk{\"u}hler, and George Drettakis.
\newblock 3d gaussian splatting for real-time radiance field rendering.
\newblock \emph{ACM Transactions on Graphics (ToG)}, 42\penalty0 (4):\penalty0 1--14, 2023.

\bibitem[Kong et~al.(2023)Kong, Liu, Taher, and Davison]{task_vmap23}
Xin Kong, Shikun Liu, Marwan Taher, and Andrew~J Davison.
\newblock vmap: Vectorised object mapping for neural field slam.
\newblock In \emph{Proceedings of the IEEE/CVF Conference on Computer Vision and Pattern Recognition}, pages 952--961, 2023.

\bibitem[Leutenegger et~al.(2015)Leutenegger, Lynen, Bosse, Siegwart, and Furgale]{tra_Leutenegger2015KeyframebasedVO}
Stefan Leutenegger, Simon Lynen, Michael Bosse, Roland~Y. Siegwart, and Paul~Timothy Furgale.
\newblock Keyframe-based visual–inertial odometry using nonlinear optimization.
\newblock \emph{The International Journal of Robotics Research}, 34:\penalty0 314 -- 334, 2015.

\bibitem[Li et~al.(2022)Li, Gu, Yuan, Dong, Tan, et~al.]{dimslam23}
Heng Li, Xiaodong Gu, Weihao Yuan, Zilong Dong, Ping Tan, et~al.
\newblock Dense rgb slam with neural implicit maps.
\newblock In \emph{The Eleventh International Conference on Learning Representations}, 2022.

\bibitem[Li et~al.(2023)Li, He, Wang, and Wang]{multi-mlp23}
Mingrui Li, Jiaming He, Yangyang Wang, and Hongyu Wang.
\newblock End-to-end rgb-d slam with multi-mlps dense neural implicit representations.
\newblock \emph{IEEE Robotics and Automation Letters}, 2023.

\bibitem[Lin et~al.(2021)Lin, Ma, Torralba, and Lucey]{pose2}
Chen-Hsuan Lin, Wei-Chiu Ma, Antonio Torralba, and Simon Lucey.
\newblock Barf: Bundle-adjusting neural radiance fields.
\newblock In \emph{Proceedings of the IEEE/CVF International Conference on Computer Vision}, pages 5741--5751, 2021.

\bibitem[Liu et~al.(2017)Liu, Li, Chen, Zhang, Kaess, and Bao]{RGBDSLAM3}
Haomin Liu, Chen Li, Guojun Chen, Guofeng Zhang, Michael Kaess, and Hujun Bao.
\newblock Robust keyframe-based dense slam with an rgb-d camera.
\newblock \emph{ArXiv}, abs/1711.05166, 2017.

\bibitem[Liu and Chen(2023)]{rim23}
Jianheng Liu and Haoyao Chen.
\newblock Towards real-time scalable dense mapping using robot-centric implicit representation.
\newblock \emph{arXiv preprint arXiv:2306.10472}, 2023.

\bibitem[Liu and Zhu(2023{\natexlab{a}})]{eMLP23}
Shaofan Liu and Jianke Zhu.
\newblock Efficient map fusion for multiple implicit slam agents.
\newblock \emph{IEEE Transactions on Intelligent Vehicles}, 2023{\natexlab{a}}.

\bibitem[Liu and Zhu(2023{\natexlab{b}})]{sbumap_map-fusion23}
Shaofan Liu and Jianke Zhu.
\newblock Efficient map fusion for multiple implicit slam agents.
\newblock \emph{IEEE Transactions on Intelligent Vehicles}, 2023{\natexlab{b}}.

\bibitem[Matsuki et~al.(2023)Matsuki, Tateno, Niemeyer, and Tombari]{newton23}
Hidenobu Matsuki, Keisuke Tateno, Michael Niemeyer, and Federic Tombari.
\newblock Newton: Neural view-centric mapping for on-the-fly large-scale slam.
\newblock \emph{arXiv preprint arXiv:2303.13654}, 2023.

\bibitem[Mildenhall et~al.(2021)Mildenhall, Srinivasan, Tancik, Barron, Ramamoorthi, and Ng]{nerf21}
Ben Mildenhall, Pratul~P Srinivasan, Matthew Tancik, Jonathan~T Barron, Ravi Ramamoorthi, and Ren Ng.
\newblock Nerf: Representing scenes as neural radiance fields for view synthesis.
\newblock \emph{Communications of the ACM}, 65\penalty0 (1):\penalty0 99--106, 2021.

\bibitem[M{\"u}ller et~al.(2022)M{\"u}ller, Evans, Schied, and Keller]{instant-npg22}
Thomas M{\"u}ller, Alex Evans, Christoph Schied, and Alexander Keller.
\newblock Instant neural graphics primitives with a multiresolution hash encoding.
\newblock \emph{ACM Transactions on Graphics (ToG)}, 41\penalty0 (4):\penalty0 1--15, 2022.

\bibitem[Mur-Artal and Tard{\'o}s(2016)]{tra_MurArtal2016ORBSLAM2AO}
Raul Mur-Artal and Juan~D. Tard{\'o}s.
\newblock Orb-slam2: An open-source slam system for monocular, stereo, and rgb-d cameras.
\newblock \emph{IEEE Transactions on Robotics}, 33:\penalty0 1255--1262, 2016.

\bibitem[Mur-Artal et~al.(2015)Mur-Artal, Montiel, and Tard{\'o}s]{tra_MurArtal2015ORBSLAMAV}
Raul Mur-Artal, Jos{\'e} M.~M. Montiel, and Juan~D. Tard{\'o}s.
\newblock Orb-slam: A versatile and accurate monocular slam system.
\newblock \emph{IEEE Transactions on Robotics}, 31:\penalty0 1147--1163, 2015.

\bibitem[Newcombe et~al.(2011)Newcombe, Izadi, Hilliges, Molyneaux, Kim, Davison, Kohli, Shotton, Hodges, and Fitzgibbon]{RGBDSLAM5}
Richard~A. Newcombe, Shahram Izadi, Otmar Hilliges, David Molyneaux, David Kim, Andrew~J. Davison, Pushmeet Kohli, Jamie Shotton, Steve Hodges, and Andrew~William Fitzgibbon.
\newblock Kinectfusion: Real-time dense surface mapping and tracking.
\newblock \emph{2011 10th IEEE International Symposium on Mixed and Augmented Reality}, pages 127--136, 2011.

\bibitem[Oechsle et~al.(2021)Oechsle, Peng, and Geiger]{UNISURFUN21}
Michael Oechsle, Songyou Peng, and Andreas Geiger.
\newblock Unisurf: Unifying neural implicit surfaces and radiance fields for multi-view reconstruction.
\newblock \emph{2021 IEEE/CVF International Conference on Computer Vision (ICCV)}, pages 5569--5579, 2021.

\bibitem[Rosinol et~al.(2022)Rosinol, Leonard, and Carlone]{nerfslam22}
Antoni Rosinol, John~J Leonard, and Luca Carlone.
\newblock Nerf-slam: Real-time dense monocular slam with neural radiance fields.
\newblock \emph{arXiv preprint arXiv:2210.13641}, 2022.

\bibitem[Sandstr{\"o}m et~al.(2023)Sandstr{\"o}m, Li, Van~Gool, and Oswald]{pointslam24}
Erik Sandstr{\"o}m, Yue Li, Luc Van~Gool, and Martin~R Oswald.
\newblock Point-slam: Dense neural point cloud-based slam.
\newblock In \emph{Proceedings of the IEEE/CVF International Conference on Computer Vision}, pages 18433--18444, 2023.

\bibitem[Sch{\"o}ps et~al.(2019)Sch{\"o}ps, Sattler, and Pollefeys]{RGBDSLAM1}
Thomas Sch{\"o}ps, Torsten Sattler, and Marc Pollefeys.
\newblock Bad slam: Bundle adjusted direct rgb-d slam.
\newblock \emph{2019 IEEE/CVF Conference on Computer Vision and Pattern Recognition (CVPR)}, pages 134--144, 2019.

\bibitem[Straub et~al.(2019)Straub, Whelan, Ma, Chen, Wijmans, Green, Engel, Mur-Artal, Ren, Verma, Clarkson, Yan, Budge, Yan, Pan, Yon, Zou, Leon, Carter, Briales, Gillingham, Mueggler, Pesqueira, Savva, Batra, Strasdat, Nardi, Goesele, Lovegrove, and Newcombe]{replica19arxiv}
Julian Straub, Thomas Whelan, Lingni Ma, Yufan Chen, Erik Wijmans, Simon Green, Jakob~J. Engel, Raul Mur-Artal, Carl Ren, Shobhit Verma, Anton Clarkson, Mingfei Yan, Brian Budge, Yajie Yan, Xiaqing Pan, June Yon, Yuyang Zou, Kimberly Leon, Nigel Carter, Jesus Briales, Tyler Gillingham, Elias Mueggler, Luis Pesqueira, Manolis Savva, Dhruv Batra, Hauke~M. Strasdat, Renzo~De Nardi, Michael Goesele, Steven Lovegrove, and Richard Newcombe.
\newblock The {R}eplica dataset: A digital replica of indoor spaces.
\newblock \emph{arXiv preprint arXiv:1906.05797}, 2019.

\bibitem[Sucar et~al.(2021)Sucar, Liu, Ortiz, and Davison]{imap22}
Edgar Sucar, Shikun Liu, Joseph Ortiz, and Andrew~J. Davison.
\newblock imap: Implicit mapping and positioning in real-time.
\newblock In \emph{2021 {IEEE/CVF} International Conference on Computer Vision, {ICCV} 2021, Montreal, QC, Canada, October 10-17, 2021}, pages 6209--6218. {IEEE}, 2021.

\bibitem[Tang et~al.(2023)Tang, Zhang, Yu, Wang, and Xu]{sbumap_mips23}
Yijie Tang, Jiazhao Zhang, Zhinan Yu, He Wang, and Kai Xu.
\newblock Mips-fusion: Multi-implicit-submaps for scalable and robust online neural rgb-d reconstruction.
\newblock \emph{arXiv preprint arXiv:2308.08741}, 2023.

\bibitem[Wang et~al.(2023)Wang, Wang, and Agapito]{coslam23}
Hengyi Wang, Jingwen Wang, and Lourdes Agapito.
\newblock Co-slam: Joint coordinate and sparse parametric encodings for neural real-time slam.
\newblock In \emph{Proceedings of the IEEE/CVF Conference on Computer Vision and Pattern Recognition}, pages 13293--13302, 2023.

\bibitem[Wang et~al.(2021)Wang, Liu, Liu, Theobalt, Komura, and Wang]{neus21}
Peng Wang, Lingjie Liu, Yuan Liu, Christian Theobalt, Taku Komura, and Wenping Wang.
\newblock Neus: Learning neural implicit surfaces by volume rendering for multi-view reconstruction.
\newblock \emph{NeurIPS}, 2021.

\bibitem[Whelan et~al.(2015)Whelan, Kaess, Johannsson, Fallon, Leonard, and McDonald]{RGBDSLAM7}
Thomas Whelan, Michael Kaess, Hordur Johannsson, Maurice Fallon, John~J Leonard, and John McDonald.
\newblock Real-time large-scale dense rgb-d slam with volumetric fusion.
\newblock \emph{The International Journal of Robotics Research}, 34\penalty0 (4-5):\penalty0 598--626, 2015.

\bibitem[Whelan et~al.(2016)Whelan, Salas-Moreno, Glocker, Davison, and Leutenegger]{RGBDSLAM4}
Thomas Whelan, Renato~F. Salas-Moreno, Ben Glocker, Andrew~J. Davison, and Stefan Leutenegger.
\newblock Elasticfusion: Real-time dense slam and light source estimation.
\newblock \emph{The International Journal of Robotics Research}, 35:\penalty0 1697 -- 1716, 2016.

\bibitem[Xiang et~al.(2023)Xiang, Sun, Xie, Yang, and Wang]{sbumap_nisb23}
Beichen Xiang, Yuxin Sun, Zhongqu Xie, Xiaolong Yang, and Yulin Wang.
\newblock Nisb-map: Scalable mapping with neural implicit spatial block.
\newblock \emph{IEEE Robotics and Automation Letters}, 2023.

\bibitem[Xu et~al.(2022)Xu, Xu, Philip, Bi, Shu, Sunkavalli, and Neumann]{pointnerf22}
Qiangeng Xu, Zexiang Xu, Julien Philip, Sai Bi, Zhixin Shu, Kalyan Sunkavalli, and Ulrich Neumann.
\newblock Point-nerf: Point-based neural radiance fields.
\newblock In \emph{Proceedings of the IEEE/CVF Conference on Computer Vision and Pattern Recognition}, pages 5438--5448, 2022.

\bibitem[Yan et~al.(2023)Yan, Qu, Wang, Xu, Wang, Zhao, and Li]{gsslam23}
Chi Yan, Delin Qu, Dong Wang, Dan Xu, Zhigang Wang, Bin Zhao, and Xuelong Li.
\newblock Gs-slam: Dense visual slam with 3d gaussian splatting.
\newblock \emph{arXiv preprint arXiv:2311.11700}, 2023.

\bibitem[Yan et~al.(2017)Yan, Ye, and Ren]{RGBDSLAM2}
Zhixin Yan, Mao Ye, and Liu Ren.
\newblock Dense visual slam with probabilistic surfel map.
\newblock \emph{IEEE Transactions on Visualization and Computer Graphics}, 23:\penalty0 2389--2398, 2017.

\bibitem[Yang et~al.(2022)Yang, Li, Zhai, Ming, Liu, and Zhang]{voxfusion22}
Xingrui Yang, Hai Li, Hongjia Zhai, Yuhang Ming, Yuqian Liu, and Guofeng Zhang.
\newblock Vox-fusion: Dense tracking and mapping with voxel-based neural implicit representation.
\newblock In \emph{2022 IEEE International Symposium on Mixed and Augmented Reality (ISMAR)}, pages 499--507. IEEE, 2022.

\bibitem[Yariv et~al.(2021)Yariv, Gu, Kasten, and Lipman]{VolumeRO21}
Lior Yariv, Jiatao Gu, Yoni Kasten, and Yaron Lipman.
\newblock Volume rendering of neural implicit surfaces.
\newblock \emph{ArXiv}, abs/2106.12052, 2021.

\bibitem[Yen-Chen et~al.(2021)Yen-Chen, Florence, Barron, Rodriguez, Isola, and Lin]{pose0}
Lin Yen-Chen, Pete Florence, Jonathan~T Barron, Alberto Rodriguez, Phillip Isola, and Tsung-Yi Lin.
\newblock inerf: Inverting neural radiance fields for pose estimation.
\newblock In \emph{2021 IEEE/RSJ International Conference on Intelligent Robots and Systems (IROS)}, pages 1323--1330. IEEE, 2021.

\bibitem[Yuan and Nuechter(2023)]{task_unifusion23}
Yijun Yuan and Andreas Nuechter.
\newblock Uni-fusion: Universal continuous mapping.
\newblock \emph{arXiv preprint arXiv:2303.12678}, 2023.

\bibitem[Zhang et~al.(2023{\natexlab{a}})Zhang, Sun, Wang, Cheng, and Haala]{hislam23arXiv}
Wei Zhang, Tiecheng Sun, Sen Wang, Qing Cheng, and Norbert Haala.
\newblock Hi-slam: Monocular real-time dense mapping with hybrid implicit fields.
\newblock \emph{arXiv preprint arXiv:2310.04787}, 2023{\natexlab{a}}.

\bibitem[Zhang et~al.(2023{\natexlab{b}})Zhang, Hu, Wu, Zhao, Li, Zou, and Fan]{towards23}
Yongqiang Zhang, Zhipeng Hu, Haoqian Wu, Minda Zhao, Lincheng Li, Zhengxia Zou, and Changjie Fan.
\newblock Towards unbiased volume rendering of neural implicit surfaces with geometry priors.
\newblock In \emph{Proceedings of the IEEE/CVF Conference on Computer Vision and Pattern Recognition}, pages 4359--4368, 2023{\natexlab{b}}.

\bibitem[Zhang et~al.(2023{\natexlab{c}})Zhang, Tosi, Mattoccia, and Poggi]{goslam23}
Youmin Zhang, Fabio Tosi, Stefano Mattoccia, and Matteo Poggi.
\newblock Go-slam: Global optimization for consistent 3d instant reconstruction.
\newblock In \emph{Proceedings of the IEEE/CVF International Conference on Computer Vision}, pages 3727--3737, 2023{\natexlab{c}}.

\bibitem[Zhong et~al.(2023)Zhong, Pan, Behley, and Stachniss]{shine-map23}
Xingguang Zhong, Yue Pan, Jens Behley, and Cyrill Stachniss.
\newblock Shine-mapping: Large-scale 3d mapping using sparse hierarchical implicit neural representations.
\newblock In \emph{2023 IEEE International Conference on Robotics and Automation (ICRA)}, pages 8371--8377. IEEE, 2023.

\bibitem[Zhu et~al.(2022)Zhu, Peng, Larsson, Xu, Bao, Cui, Oswald, and Pollefeys]{niceslam22}
Zihan Zhu, Songyou Peng, Viktor Larsson, Weiwei Xu, Hujun Bao, Zhaopeng Cui, Martin~R. Oswald, and Marc Pollefeys.
\newblock {NICE-SLAM:} neural implicit scalable encoding for {SLAM}.
\newblock In \emph{{IEEE/CVF} Conference on Computer Vision and Pattern Recognition, {CVPR} 2022, New Orleans, LA, USA, June 18-24, 2022}, pages 12776--12786. {IEEE}, 2022.

\bibitem[Zhu et~al.(2023)Zhu, Peng, Larsson, Cui, Oswald, Geiger, and Pollefeys]{nicer23}
Zihan Zhu, Songyou Peng, Viktor Larsson, Zhaopeng Cui, Martin~R Oswald, Andreas Geiger, and Marc Pollefeys.
\newblock Nicer-slam: Neural implicit scene encoding for rgb slam.
\newblock \emph{arXiv preprint arXiv:2302.03594}, 2023.

\end{thebibliography}
}

\clearpage
\setcounter{page}{1}
\maketitlesupplementary

\section{Implementation Details}
\label{sec:implement}

We conducted the benchmark leaderboard using a 2.60GHz Intel Xeon Platinum 8358P CPU and an A800-SXM4-80GB GPU. In contrast, the influence of network structure (refer to Table 1 in the main paper) and mapping time of Explicit Hybrid Encoding was evaluated on a 2.10GHz Intel Xeon Gold 5218R CPU and an NVIDIA GeForce RTX 3090.
Our experiments were carried out within the unified SLAM framework, utilizing synthetic datasets for consistency and control. The system initialization was performed with the first posed RGB-D frame, followed by an optimization over 500 iterations. For subsequent tracking and mapping, we conducted optimizations over 20 iterations, sampling 1024 and 2048 pixels, respectively. A keyframe was defined as every 5\textsuperscript{th} input frame. Here, 5\% of the pixels were randomly sampled and stored for global bundle adjustment purposes. For each sampled pixel, we selected 60 coordinates along its ray. Within the truncated region ($T=10$cm), 12 coordinates were uniformly sampled, and an additional 48 were drawn uniformly from the free space for pixels with depth information. In cases without depth information, the 60 points were uniformly sampled. 
All hybrid representations (\ie $\mathcal{F} !=$ MLP) utilized coarse and fine feature spatial splittings with resolutions of 24cm and 2cm, respectively. The feature dimensions for each resolution level were set to 2 and concatenated into a 4-dimensional vector for processing by color and geometry decoders. These decoders are comprised of small, shallow MLPs with 2 layers and 32 hidden channels. We configured the color loss as $\lambda_p=5$, the rendered depth loss as $\lambda_g=1$, and the SDF losses as $\lambda_t=200$, $\lambda_c=50$, and $\lambda_{fs}=10$, respectively.

\begin{figure}[t!]
  \centering
   \includegraphics[width=1\linewidth]{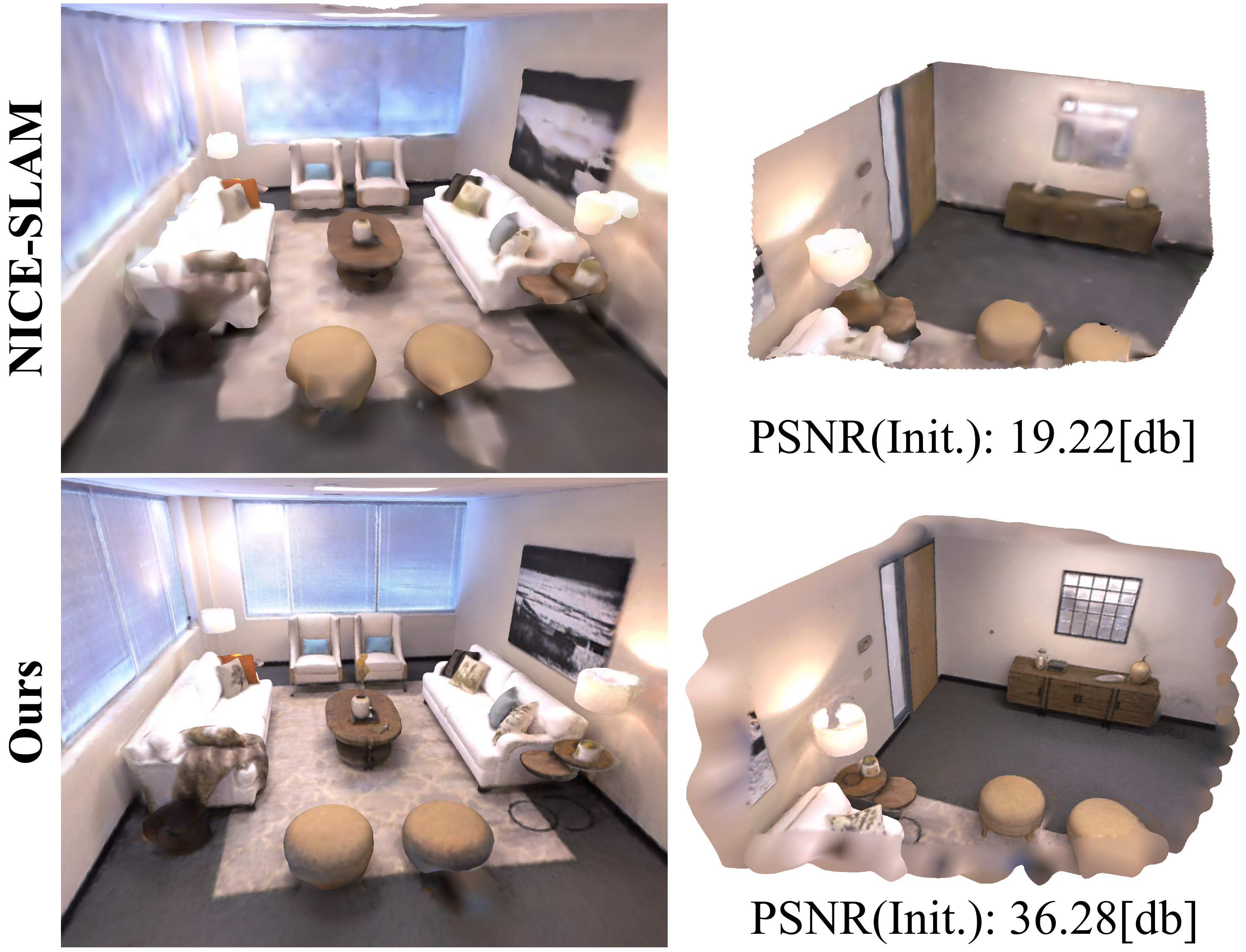}
   \vspace{-10pt}
   \caption{Comparison of the hierarchical dense grid as implemented in our benchmark with the previous state-of-the-art (SOTA)~\cite{niceslam22}. Right column: Meshes trained on the initial RGB-D frame for 1000 iterations.}
   \label{fig:cover}
   \vspace{-10pt}
\end{figure}

\begin{figure}[t!]
  \centering
   \includegraphics[width=1\linewidth]{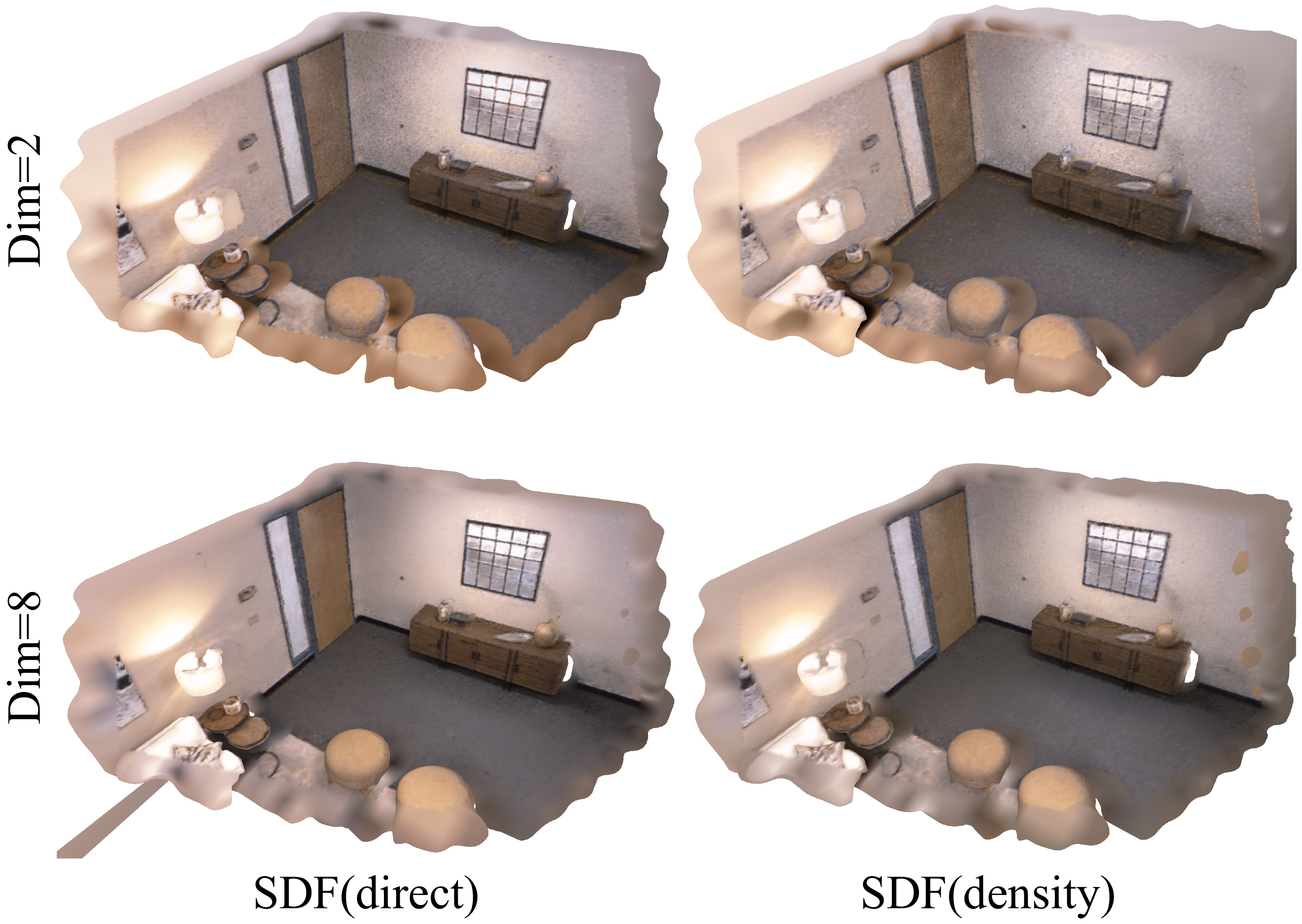}
   \vspace{-10pt}
   \caption{Dense grid initialization visualized at the first posed frame over 500 iterations for different rendering functions, featuring 2 and 8-dimensional settings.}
   \label{fig:effect_dim}
   \vspace{-10pt}
\end{figure}

\noindent\textbf{Cover figure} (\cref{fig:cover}) was produced using configurations that differ slightly from our standard setup. For this figure, the initialization phase was extended to 1000 iterations. We documented the highest \textit{PSNR} value obtained in the last three iterations for both our method and NICE-SLAM. For mapping, we increased the number of sampled pixels to 4080, while for tracking purposes, we used 2048 pixels. The iterations for mapping and tracking were set to 30 and 20, respectively. Notably, the figure relies on SDF (density) functions with 8-dimensional features, which is $\times$4 less than the configuration used in~\cite{niceslam22} and does not require pre-trained models. 

\begin{figure}[t!]
  \centering
   \includegraphics[width=1\linewidth]{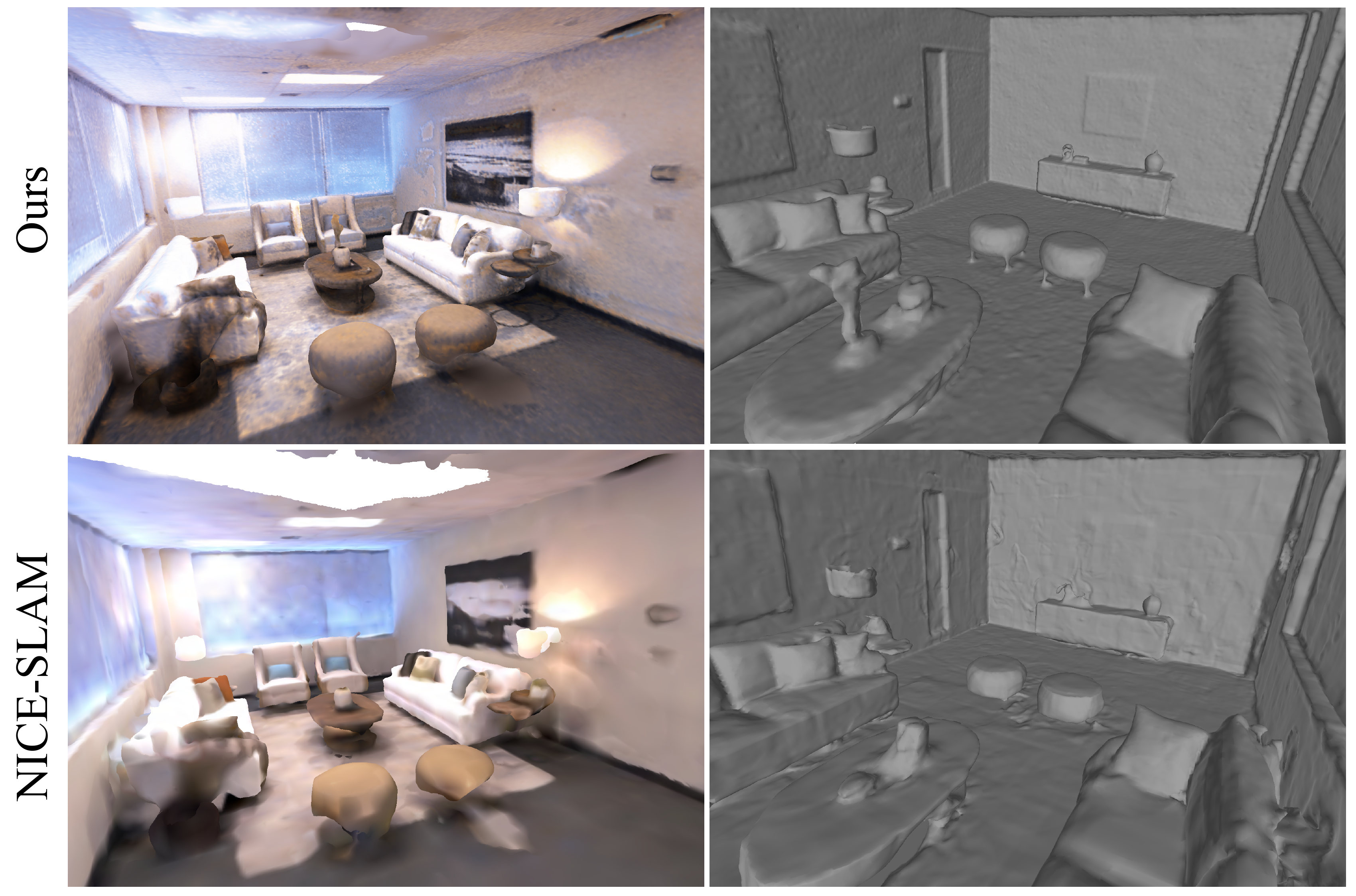}
   \caption{Dense Grid (density) compared to NICE-SLAM under benchmark configurations (Dimensions$\times$Resolutions are 2$\times$2 for Ours vs 32$\times$3 for NICE-SLAM).}
   \label{fig:cover_baseline}
\end{figure}

The differences in feature dimensions and SDF transformation functions utilized during initialization are further detailed in~\cref{fig:effect_dim}. The reconstructed accuracy (\textit{Acc.}) for~\cref{fig:cover} is reported as 1.53cm, with \textit{ATE} of 0.73cm. This represents a significant improvement, approximately 60\% better, compared to the 3.87cm (\textit{Acc.}) and 1.95cm (\textit{ATE}) achieved by NICE-SLAM. The \textit{ATE} record for NICE-SLAM can be found in Table 1 of~\cite{voxfusion22}.
A comparative demonstration of the dense grid representation with our baseline configuration, used for the main leaderboard, can be found in~\cref{fig:cover_baseline}.

\noindent\textbf{Training specification.} Firstly, we specify the photometric and geometric losses in accordance with Eq. (4) in the main paper:
{\setlength\abovedisplayskip{2pt}
\setlength\belowdisplayskip{2pt}
\begin{equation}
    \mathcal{L}_p =\frac{1}{\mathcal{M}} \sum\limits_{x} ({e^p_x})^2, \space \\
\mathcal{L}_g =\frac{1}{\mathcal{M}_{sub}} \sum\limits_{x} ({e^g_x})^2
\label{eq:loss_basic}
\end{equation}}In this context, \(\mathcal{M}\) denotes the global list of sampling pixels (\eg, key frames), and \(\mathcal{M}_{\text{sub}}\) represents the subset of those pixels with non-zero depth values. To estimate the SDF \( s_i \) for each sample point \( p_i \), a geometric constraint is applied:
{\setlength\abovedisplayskip{2pt}
\setlength\belowdisplayskip{2pt}
\begin{equation}
e^{sdf}_i= d_i +\tilde{s}_i \cdot T - d_x
\label{eq:loss_sdf_e}
\end{equation}}Hence, for points \( p_i \) located within the truncated region \( i \in \mathcal{P}^T \), defined by the condition \( |d_x - d_i| < T \), the SDF loss is formulated as follows:
{\setlength\abovedisplayskip{2pt}
\setlength\belowdisplayskip{2pt}
\begin{equation}
\mathcal{L}_{sdf} =\frac{1}{\mathcal{M}_{sub}} \sum\limits_{x} \frac{1}{\mathcal{P}^T} \sum\limits_{i}({e^{sdf}_i})^2
\label{eq:loss_sdf_e}
\end{equation}}Moreover, by defining the area between the camera center and the truncation boundary \( T \) as free space, we apply constraints to points in this region, denoted as \( i \in \mathcal{P}^{\text{fs}} \):
{\setlength\abovedisplayskip{2pt}
\setlength\belowdisplayskip{2pt}
\begin{equation}
e^{fs}_x= \tilde{s}_x - 1
\label{eq:loss_fs_e}
\end{equation}}This leads to the establishment of the free space loss:
{\setlength\abovedisplayskip{2pt}
\setlength\belowdisplayskip{2pt}
\begin{equation}
\mathcal{L}_{fs} = \mathcal{L}_{fs} =\frac{1}{\mathcal{M}_{sub}} \sum\limits_{x} \frac{1}{\mathcal{P}^{fs}} \sum\limits_{i}({e^{fs}_i})^2
\label{eq:loss_fs}
\end{equation}}Following the approach described in~\cite{eslam23}, we differentiate the importance of samples within the truncation region by applying distinct weighting factors: \(\lambda_c\) for samples close to the surface and \(\lambda_t\) for those near the truncation boundary. The overall objective function is as follows:
{\setlength\abovedisplayskip{2pt}
\setlength\belowdisplayskip{2pt}
\begin{equation}
\mathcal{L} = \lambda_p \mathcal{L}_p + \lambda_g \mathcal{L}_g + \lambda_t \mathcal{L}_{sdf}^t + \lambda_c \mathcal{L}_{sdf}^c + \lambda_{fs} \mathcal{L}_{fs}
\label{eq:total_loss}
\end{equation}}

\section{Performance Trade-off}
\label{sec:tradeoff}

\begin{figure}[t!]
  \centering
     \includegraphics[width=1\linewidth]{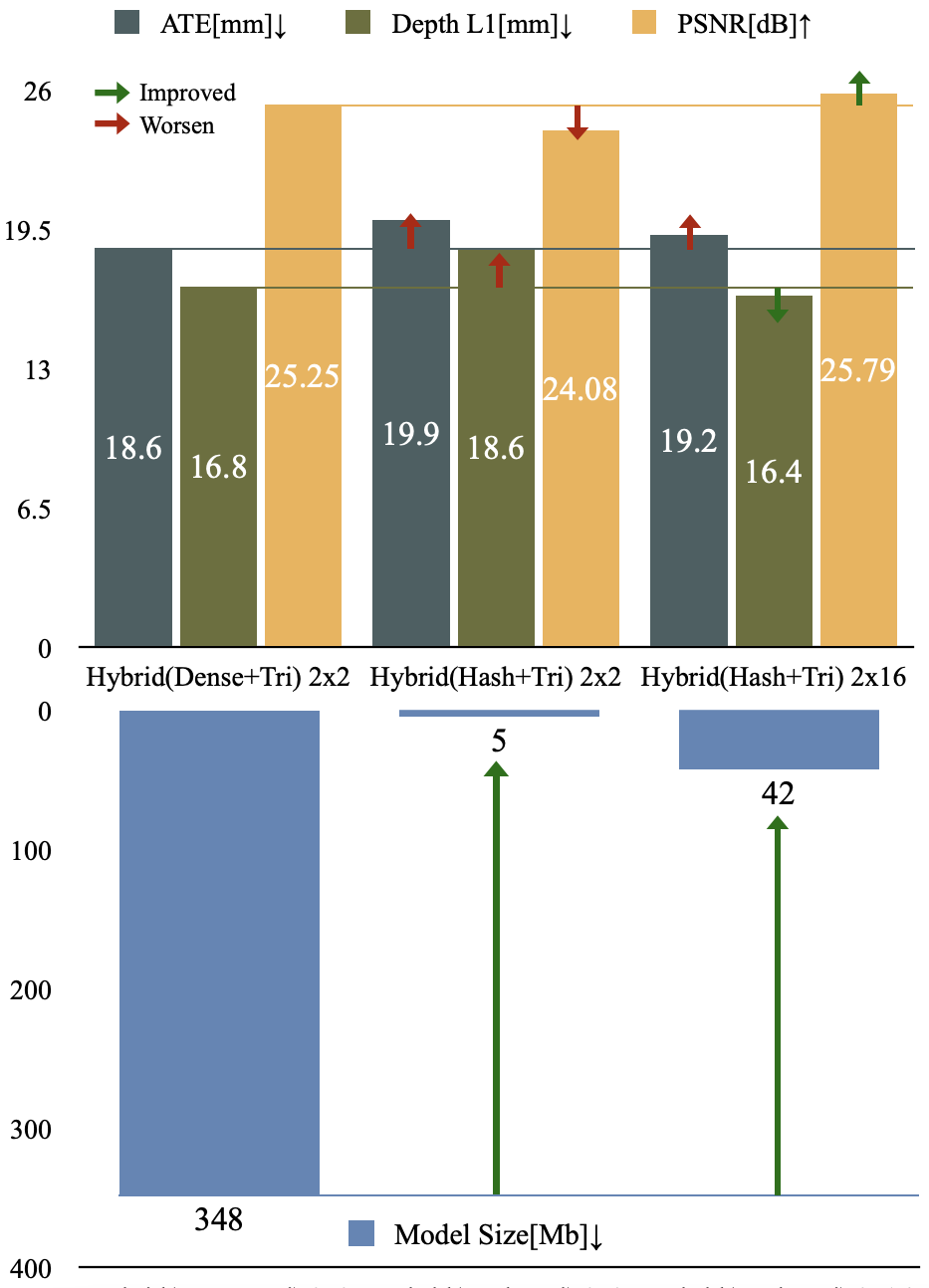}
     \vspace{-10pt}
   \caption{Memory-performance trade-offs: Using dense grid-based hybrid encoding as a baseline, hash-based alternatives demonstrate significant memory savings. This efficiency is maintained even with an $\times8$ increase in resolution levels for hybrid encoding of hash grid and tri-plane.}
   \label{fig:tradeoff}
\end{figure}

\textbf{Benchmark Considerations.} Our benchmark study strictly controls feature dimensions and spatial resolution. However, optimal performance in such controlled comparisons may not always align with real-world application priorities. For example, a minor compromise in accuracy (a 1mm decrease in trajectory estimation accuracy) can lead to substantial gains in efficiency, such as an 87.93\% reduction in memory usage, as shown in~\cref{fig:tradeoff}. Hence, in practical scenarios, \textit{a hybrid approach combining tri-plane and hash encodings} might be a preferable alternative to the tri-plane and dense grid combination.

\begin{figure*}[t!]
  \centering
   \includegraphics[width=1\linewidth]{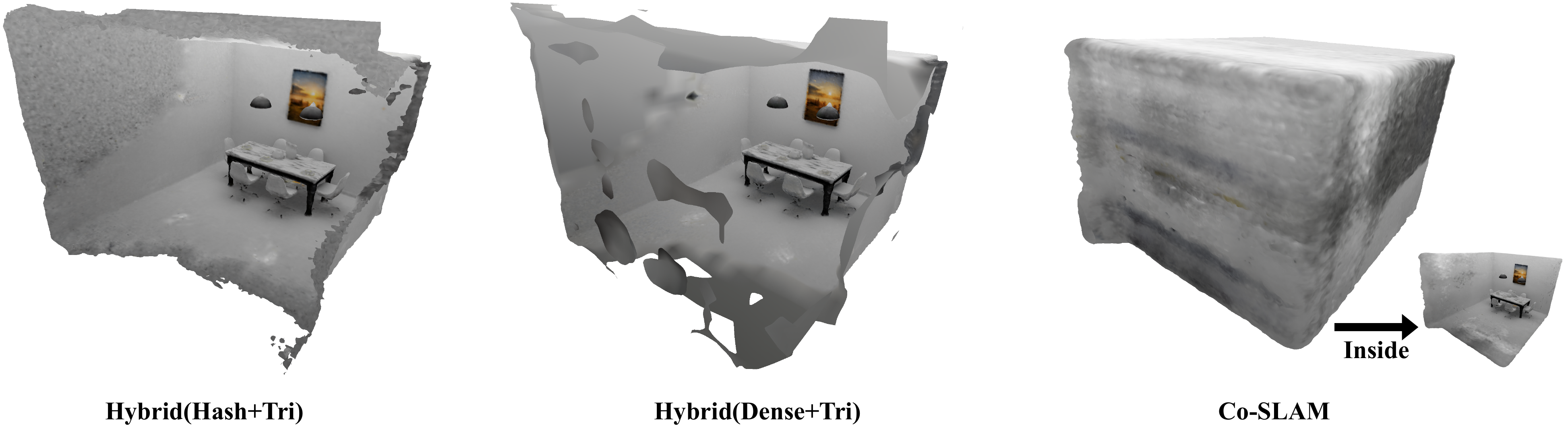}
   \vspace{-10pt}
   \caption{Visualization of \textit{'breakfast room'} sequence of NeuralRGBD dataset~\cite{neuralrgbd22}.}
   \label{fig:arti}
   \vspace{-10pt}
\end{figure*}

\noindent\textbf{Comparison with Existing Work.} 
We compared our hybrid designs with a SLAM system that employs similar joint encoding strategies, \ie Co-SLAM~\cite{coslam23}. The dense grid-based hybrid encoding exhibits superior performance, albeit at the cost of significantly higher memory consumption~\cref{fig:tradeoff}. Interestingly, the Hybrid (Hash+Tri) configuration, while not excelling in inference capabilities or memory efficiency in~\cref{tab:compute}, emerges as the most practical option for real-world applications. As illustrated in~\cref{fig:arti}, our hybrid encoding notably reduces artifacts compared to Co-SLAM. This is achieved with a minimal trade-off in trajectory accuracy (only 1mm lower than Co-SLAM) for the Hash+Tri combination. Regarding inferring performance, our Hybrid (Dense+Tri) encoding achieves the best \textit{ATE}, \textit{Depth L1}, and \textit{PSNR} scores, as detailed in~\cref{tab:compute}.

\tabcolsep=0.07cm
\begin{table}[t!]
\small
\centering
\begin{tabular}{ccccc}
\toprule
SLAM
&\makecell[c]{ATE$\downarrow$ \\ {[}cm{]}} 
&\makecell[c]{Depth L1$\downarrow$ \\{[}cm{]}} 
&\makecell[c]{PSNR$\uparrow$ \\{[}dB{]}} 
&\makecell[c]{Model Size$\downarrow$ \\{[}MB{]}}  \\
\hline
\makecell[c]{Hybrid(Dense+Tri)\\ $2\times8$}
                    &\textbf{1.73}  &\textbf{1.55} &\textbf{26.46}  &1364.28  \\
\makecell[c]{Hybrid(Hash+Tri)\\ $16\times2$}
                    &1.92  &1.64 &25.79  &42.45  \\
\makecell[c]{Co-SLAM*\\ $16\times2$}
                    &1.83  &1.64  &25.41  &\textbf{7.25} \\
\bottomrule
\end{tabular}
\vspace{-5pt}
\captionsetup{justification=justified}
\caption{Grid-based Hybrid Representations vs. Co-SLAM (\cite{coslam23}). The asterisk denotes Co-SLAM's performance as measured by implementing its open-source code.}
\label{tab:compute}
\vspace{-5pt}
\end{table}

\tabcolsep=0.07cm
\begin{table}[t!]
\small
\centering
\begin{tabular}{cccc}
\toprule
SLAM
&Acc.$\downarrow$ {[}cm{]}
&Comp.$\downarrow${[}cm{]}
&Comp.\% $\uparrow$ {[}\%{]} \\
\hline
\makecell[c]{Hybrid(Dense+Tri)\\ $2\times2$}
                    & 2.40 & 4.64 & 83.48\\
\makecell[c]{Hybrid(Dense+Tri)\\ $2\times8$}
                    & 2.41 & 4.64 & 83.81 \\
\makecell[c]{Hybrid(Hash+Tri)\\ $16\times2$}
                    & 2.48 & 4.67 & 83.83\\
\makecell[c]{Co-SLAM*\\ $16\times2$}
                    & \textbf{2.23} & \textbf{4.52} & \textbf{84.27} \\
\bottomrule
\end{tabular}
\vspace{-5pt}
\caption{Comparison of Final Reconstruction Metrics. Evaluation based on the culling method from~\cite{neuralrgbd22}.}
\label{tab:acc}
\vspace{-10pt}
\end{table}

\noindent\textbf{Pose Estimation \& Scene Completion.} Previous research has established that greater completeness in unobserved regions often enhances overall trajectory estimation~\cite{niceslam22,coslam23}. Nonetheless, deploying the entire coarse level of the dense grid is computationally demanding~\cite{niceslam22}, and alternative one-blob encoding strategies can produce excessive artifacts~\cite{coslam23}, potentially posing risks or inconveniences in applications like robotic navigation. Our hybrid encoding, which combines a tri-plane and hash grid, successfully addresses these challenges. Ultimately, trajectory accuracy seems to be more dependent on precise scene perception, as indicated by superior \textit{PSNR} and \textit{Depth L1} metrics in~\cref{tab:compute}, than on completeness~\cref{tab:acc}.

\noindent\textbf{ About sampling \& loss \& keyframe settings.} 
Although settings like loss \& keyframe are critical and might pose different levels of impact on $\mathcal{F}$ \& $\mathcal{G}$, these are not the focus of our work. The reasons are two-fold: 
\textbf{1)} Given the intention of benchmarking, we must avoid an explosion of choices that overshadow the key insights within limited pages.
\textbf{2)} Discussion of these choices in the current benchmark might be less insightful. We suggest establishing an additional novel benchmark based on real-world challenges, e.g. large-scale and multi-agents scenarios~\cite{sbumap_mips23,cpslam23}, where settings like loss \& keyframe selection are discussed within more meaningful contexts, rather than just facilitating numerical improvement in current benchmark scenarios. 
Nevertheless, we provide a toy example of setting alternatives from other SOTA baselines on '\textit{room0}' sequence in~\cref{tab:strategy}.
\tabcolsep=0.06cm
\begin{table}[t!]
\small
\centering
\begin{tabular}{cccccc}
\toprule
Strategy  & DepthL1$\downarrow$ & PSNR$\uparrow$ & TrackTime$\downarrow$ & MapTime$\downarrow$\\ 
\hline
Sampling variant & 3.41\textcolor{green}{$\uparrow$} & 24.59\textcolor{red}{$\downarrow$} &  274\textcolor{green}{$\uparrow$}  & 752\textcolor{green}{$\uparrow$} \\
Loss variant & 85.5 \textcolor{red}{$\downarrow$} & 22.47 \textcolor{red}{$\downarrow$} &  290 \textcolor{green}{$\uparrow$} & 776 \textcolor{green}{$\uparrow$}\\
Keyframe variant& 4.82\textcolor{green}{$\uparrow$}  &26.00 \textcolor{green}{$\uparrow$}  &  279 \textcolor{green}{$\uparrow$}  & 757 \textcolor{green}{$\uparrow$} \\
\hline
Original &4.83 &25.58 & 300 & 785\\
\bottomrule
\end{tabular}
\captionsetup{justification=justified}
\caption{\textcolor{green}{$\uparrow$} \& \textcolor{red}{$\downarrow$} indicating improvement \& worsen compared to the original F(dense)+G(density). Variants are on importance sampling, free space loss, and global keyframe sampling, respectively.}
\label{tab:strategy}
\end{table}

\section{Runtime}
\label{sec:runtime}
\begin{figure*}[t!]
  \centering
   \includegraphics[width=1\linewidth]{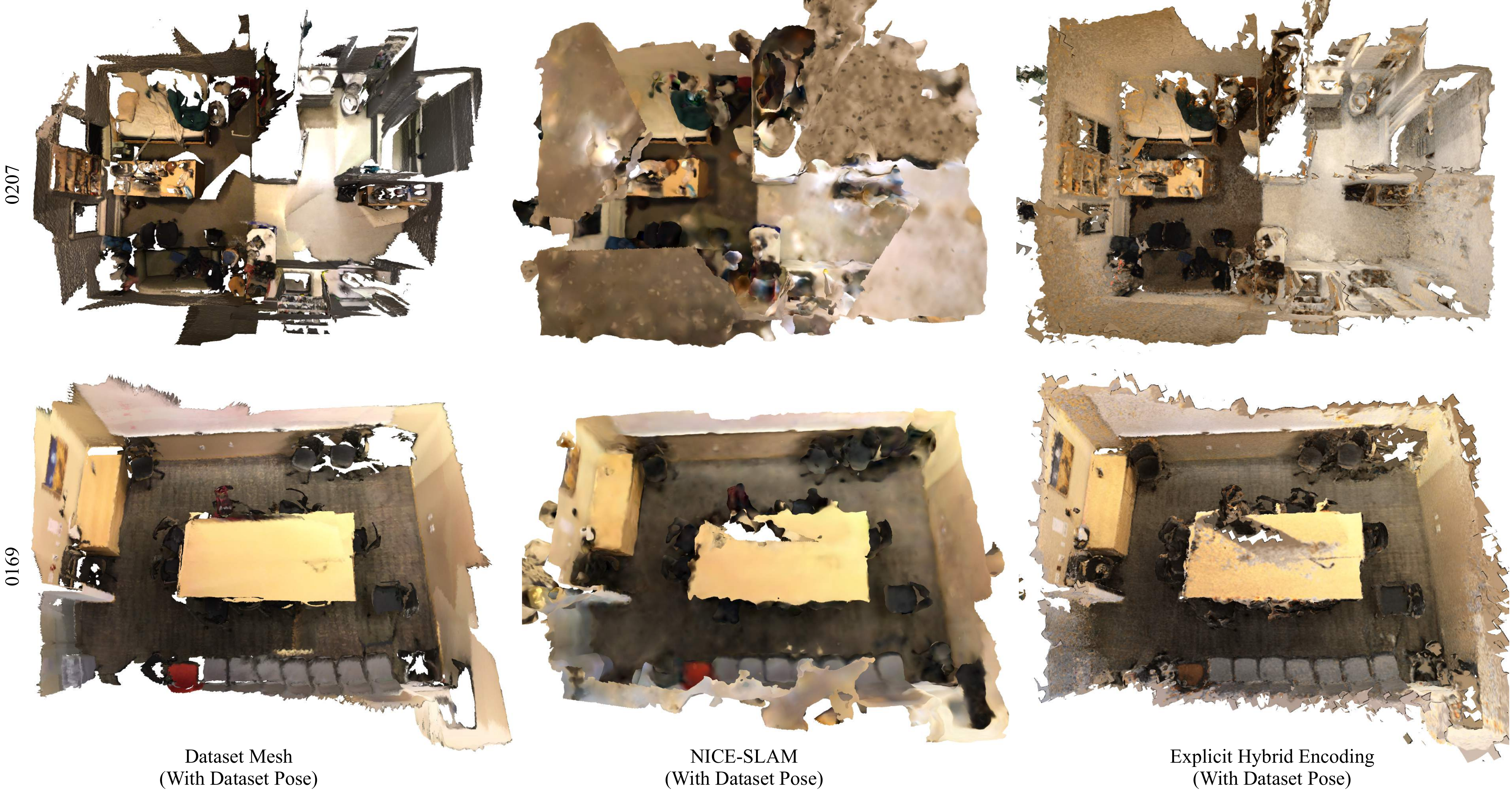}
   \caption{Qualitative evaluation of Hybrid Explicit Encoding on ScanNet Dataset. Both NICE-SLAM and Ours run on the posed RGB-D stream to simulate an externally provided tracker.}
   \label{fig:mapping}
\end{figure*}

We report the runtime of hybrid encoding (for SLAM) and explicit hybrid encoding (for mapping), evaluated on NeuralRGBD~\cite{neuralrgbd22} and Scannet Dataset~\cite{scannet17} in~\cref{tab:map_time} and~\cref{tab:hybrid_time}, respectively. For uniformly recorded map updating times, explicit hybrid encoding achieves speeds approximately $\times3$ faster than NICE-SLAM. Visual demonstration is available in~\cref{fig:mapping}.

\tabcolsep=0.07cm
\begin{table}[h!]
\small
\centering
\begin{tabular}{ccccccc}
\toprule
SLAM
&0000
&0059
&0169
&0181
&0207
&Avg.
\\
\hline
\makecell[c]{NICE-SLAM*\\ $3\times32$}
                    & 4.65 & 3.74 & 4.39 & 3.28 & 3.20 & 3.83\\
\makecell[c]{Explicit Hybrid Encoding\\ $1\times2$}
                    & \textbf{2.14} & \textbf{1.02} & \textbf{0.94} & \textbf{0.73} & \textbf{0.87} & \textbf{1.14}\\
\bottomrule
\end{tabular}
\vspace{-5pt}
\captionsetup{justification=justified}
\caption{Runtime (explicit hybrid encodings) Comparison Measured in Seconds. The asterisk denotes NICE-SLAM runtimes obtained using ground truth poses, based on its open-source code.}
\label{tab:map_time}
\end{table}

\begin{table}[h!]
\small
\centering
\begin{tabular}{cccccccccc}
\toprule
Time
& Hybrid
&br
&ck
&gr
&gwr
&ma
&tg
&w
&Avg.
\\
\hline

\multirow{2}{*}{\rotatebox{90}{Tracking}}  &\makecell[c]{Dense+Tri\\ $2\times8$}
                    & 267 & 512 & 370 & 288 & \textbf{275} & 281 & 354 & 335 \\
&\makecell[c]{Hash+Tri \\ $2\times8$}
                    & \textbf{248} & \textbf{238} & \textbf{242} & 242 & 279 & \textbf{277} & \textbf{235} & \textbf{251}\\

\hline

\multirow{2}{*}{\rotatebox{90}{Mapping}}  &\makecell[c]{Dense+Tri\\ $2\times8$}
                    & 713 & 2441 & 1502 & 903 & 676 & 654 & 1381 & 1181\\
&\makecell[c]{Hash+Tri \\ $2\times8$}
                    & \textbf{572} & \textbf{536} & \textbf{563} & \textbf{558} & \textbf{631} & \textbf{614} & \textbf{555} & \textbf{575}\\

\bottomrule
\end{tabular}
\vspace{-5pt}
\captionsetup{justification=justified}
\caption{Runtime (hybrid encodings) Comparison Measured in Milliseconds.}
\label{tab:hybrid_time}
\vspace{-10pt}
\end{table}

\section{Limitations}
\label{sec:future}

In summary, the scene representations explored in this paper primarily utilize spatial splitting along orthogonal coordinates. However, recent advancements in point-based methods~\cite{pointslam24,cpslam23,gsslam23} have demonstrated the effectiveness of newer scene representations like PointNeRF~\cite{pointnerf22} and 3D Gaussian~\cite{3dgaussian23}, offering greater adaptability. This adaptability paves the way for more sophisticated and robust systems that could integrate SLAM-centric and NeRF-centric methods, which are currently addressed separately in our work.

Point-based methods, whether modeled as 3D Gaussians with splatting-based rasterization~\cite{gsslam23} or as neural points with volume rendering~\cite{cpslam23,pointslam24}, essentially handle points in space. In neural point clouds, points directly represent spatial features, augmented by neural network processing. For 3D Gaussian representations, each point acts as the center of a Gaussian distribution, indicating not only a specific location but also a surrounding area of influence. Notably, splatting rasterization is generally less computationally demanding than volume rendering, which involves intricate integration across the volume as noted in~\cite{3dgaussian23,gsslam23}.

Looking ahead, we anticipate a more comprehensive evaluation that includes these recent approaches in implicit scene representation and geometric rendering, under a robustified SLAM system design, which incorporates essential components like loop closure and odometry, to unify both SLAM-centric and NeRF-centric methodologies.

\end{document}